# Quantifying the Knowledge Proximity Between Academic and Industry Research: An Entity and Semantic Perspective

Hongye Zhao [a], Yi Zhao [b], Chengzhi Zhang [a,*]

[a] *Department of Information Management, Nanjing University of Science and Technology, Nanjing, 210094 China*

[b] *School of Management, Anhui University, Hefei, 230601, China.*

**Abstract**: The academia and industry are characterized by a reciprocal shaping and dynamic feedback mechanism. Despite distinct institutional logics, they have adapted closely in collaborative publishing and talent mobility, demonstrating tension between institutional divergence and intensive collaboration. Existing studies on their knowledge proximity mainly rely on macro indicators such as the number of collaborative papers or patents, lacking an analysis of knowledge units in the literature. This has led to an insufficient grasp of fine-grained knowledge proximity between industry and academia, potentially undermining collaboration frameworks and resource allocation efficiency. To remedy the limitation, this study quantifies the trajectory of academia-industry co-evolution through fine-grained entities and semantic space. In the entity measurement part, we extract fine-grained knowledge entities via pre-trained models, measure sequence overlaps using cosine similarity, and analyze topological features through complex network analysis. At the semantic level, we employ unsupervised contrastive learning to quantify convergence in semantic spaces by measuring cross-institutional textual similarities. Finally, we use citation distribution patterns to examine correlations between bidirectional knowledge flows and similarity. Analysis reveals that knowledge proximity between academia and industry rises, particularly following technological change. This provides textual evidence of bidirectional adaptation in co-evolution. Additionally, academia's knowledge dominance weakens during technological paradigm shifts. Knowledge within the field exhibits a more balanced and reciprocal pattern, promoting the transformation of academia-industry relations from institutional logic divergence to knowledge production synergy. The dataset and code for this paper can be accessed at https://github.com/tinierZhao/Academic-Industrial-associations.

* Corresponding author. *E-maiL*: zhangcz@njust.edu.cn.

# 1 Introduction

As the two core subjects of the innovation ecosystem, the co-evolution between academia and industry (Blankenberg & Buenstorf, 2016; Lewin & Volberda, 1999), not only breaks through the unidirectional sequence of the traditional linear model of innovation (Bush, 1945), but also influences the scientific and technological policy research focus through the reconfiguration of the knowledge production paradigm (Elzinga, 1997; Hoppmann, 2021; Nsanzumuhire et al., 2021).

Multi-dimensional adaptations such as the unification of technical standards and the flow of innovative factors (Aufrant, 2022; Blankenberg & Buenstorf, 2016; Chai & Shih, 2016; Jiang et al., 2024) have laid the foundation for bilateral synergy. However, there are still tensions hindering in-depth cooperation (Vallas & Kleinman, 2007), including differences in research orientations (Gans et al., 2017; Merton, 1968; Teece, 1986), the divergence between academic reputation incentives and industrial competition and confidentiality mechanisms (Chirico et al., 2020; Clarysse et al., 2023; Haeussler, 2011), and cooperation frictions caused by high transaction costs (Meissner et al., 2022; Zaini et al., 2018). To address tensions in collaboration, there has been a continuous increase in specific policies aimed at optimizing academia-industry collaboration and improving the efficiency of innovation output (Hoppmann, 2021; Perkmann et al., 2013). These policies promote closer industry-academia ties and accelerate the innovation process (Bikard & Marx, 2020). This close relationship is characterized by the proximity, such as geographical proximity, and institutional proximity. They directly affect the initiation probability and frequency of academia-industry collaboration and the quality of research output (Rossi et al., 2024).

As a key dimension of proximity, knowledge (or cognitive) proximity characterizes the similarity and consistency of the knowledge bases between subjects (Brown & Duguid, 1998; Nooteboom, 2000; Rossi et al., 2024). It can both reveal the static features of their knowledge matching and dynamically track the actual effects of their collaboration (Song & Wei, 2024; Syafiandini et al., 2024). It presents unique regularities in the influence mechanisms, interaction effects, and innovation outcomes of collaboration (Nooteboom et al., 2007). However, policy promotion has limited effectiveness in practice as it fails to resolve core issues such as the misalignment of industrial and academic knowledge bases (Fischer et al., 2019). It should be noted that this study takes knowledge proximity as the core research object. Similarity refers specifically to the quantitative characterization of knowledge proximity, while interaction is the external manifestation of this proximity between industry and academia, as shown in **Fig. 1**.

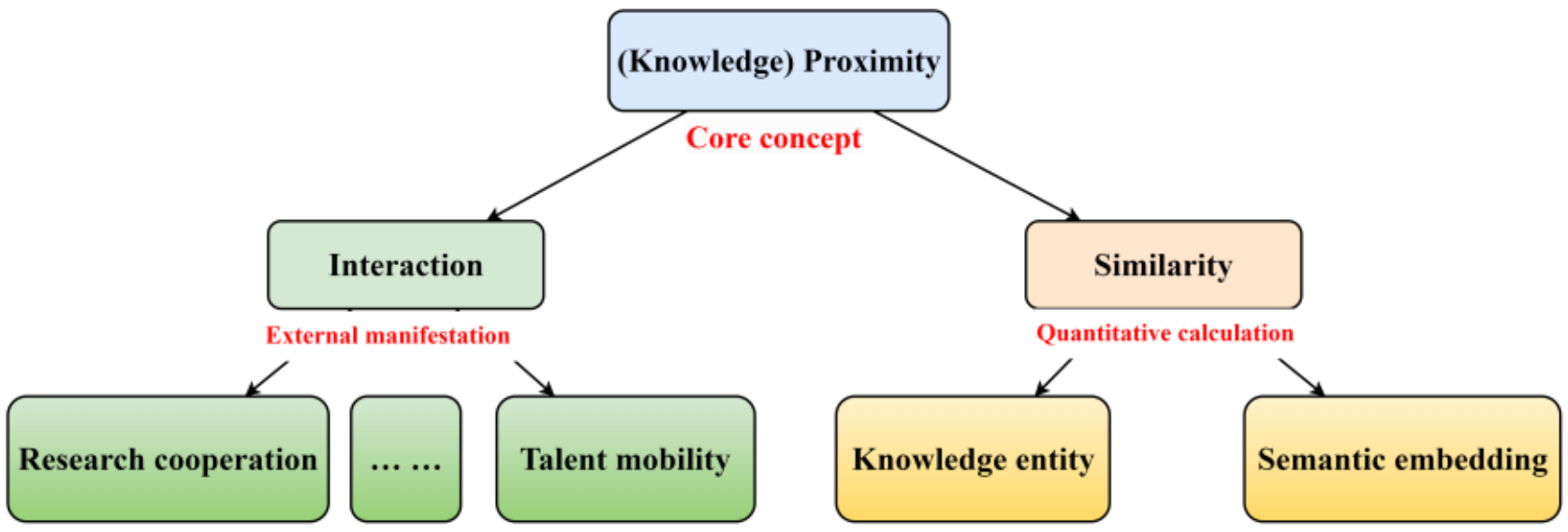

**Fig. 1.** Conceptual framework diagram of proximity, similarity, and interaction

The precise quantification of knowledge proximity has become a core lever for optimizing academia-industry collaboration models, aligning research directions, and balancing commercial value with public needs (Caviggioli et al., 2023; Hoppmann, 2021; Zaini et al., 2018). Specifically, knowledge proximity can reduce communication costs and accelerate knowledge flow (Chen, Mao, Ma, et al., 2024; Dolfsma & van der Eijk, 2016; Sudhindra et al., 2020). The gap between industry and academia can hinder knowledge dissemination and affect technological upgrading progress (Fischer et al., 2019; Hoppmann, 2021). From a social interest perspective, collaboration with close knowledge domains can better enhance the understanding of societal needs and create greater value for society (Rossi et al., 2017; Villani et al., 2017), thereby avoiding the neglect of public value under industry dominance (Ahmed et al., 2023). However, an imbalance in knowledge proximity can also trigger reverse effects. Heterogeneous knowledge is crucial for innovation performance (Rodan & Galunic, 2004). Excessively close knowledge proximity may lead to technological lock-in and path dependency within the industry (Hoppmann, 2021). Therefore, accurately grasping the knowledge proximity can both provide support for the dynamic adjustment of academia-industry collaboration strategies and offer a theoretical basis and practical guidance for the government to formulate relevant policies.

Previous studies have mainly relied on structured data to measure the knowledge proximity. For instance, Fischer et al. (2019) measured the academia-industry synergy with a patent collaboration network. Similarly, Blankenberg and Buenstorf (2016) analyzed paper and patent collaborations as well as talent mobility between industry and universities through name matching. Some scholars have utilized subclasses of the International Patent Classification (IPC) (Caviggioli et al., 2023; Colombelli et al., 2021) to calculate patent share differences across technology categories via Euclidean distances for technology convergence trends. However, IPC categories are overly broad, and the classification system updates lag technological evolution (Aceves & Evans, 2024). Liu and Zhu (2023) quantified inter-organizational cooperation stability using the $T_{uig}$ index within the Triple Helix model framework.

The bibliometric advantage of these macro indicators is the convenience of processing structured data (Puccetti et al., 2023). However, most information in scientific literature is encoded as unstructured strings (Ding et al., 2013), and traditional analyses fail to capture the detailed technical content. Although text mining techniques have supplemented traditional bibliometric indicators, existing text mining techniques for measuring knowledge linkage and similarity, including topic models and keyword extraction methods, still have limitations (Ba et al., 2024; Chen et al., 2023; Glänzel & Thijs, 2017; Liu & Zhu, 2023; Meng et al., 2024). Topic models are constrained by the subjectivity of topics and their coarse-grained representations, while keyword extraction suffers from interference from generic terms and a lack of domain specificity, making it difficult to achieve a precise characterization of technical details. This results in the unclear proximity between academia and industry in fine-grained knowledge production, which may affect the formulation of cooperation strategies between the two sectors and the government's allocation of research resources.

As the basic units and structural elements of knowledge in literature (Ding et al., 2013), knowledge entities embody specific research methods, research tools, datasets, or evaluation criteria in literature (Zhang, Zhang, et al., 2024), and serve as concrete elements for advancing the knowledge discovery process (Ding et al., 2013). Concepts and conceptual spaces form the foundation of human cognition and organizational theory (Aceves & Evans, 2024; Hofstadter & Sander, 2013). Encoding conceptual systems into multidimensional semantic spaces enables high-density representations with minimal information loss and transforms concepts and organizations into standardized geometric objects for cross-organizational comparisons (Aceves & Evans, 2024).

Based on this, this study aims to overcome the limitations of traditional bibliometric methods and existing text mining techniques through fine-grained knowledge entities and semantic space. Taking the technology evolution in the Natural Language Processing (NLP) from 2000–2022 as an empirical scenario, we focus on the following three questions:

*RQ1*: How to construct a fine-grained measurement indicator to characterize knowledge proximity, to address the limitations of existing macro-level measurement approaches?

As a key dimension for analyzing academia-industry synergy, its value lies not only in identifying static conclusions about the similarity between the knowledge bases of both parties but also in revealing the dynamic patterns of proximity between industry and academia (Caviggioli et al., 2023; Fischer et al., 2019; Hoppmann, 2021).

*RQ2*: How does the knowledge proximity between academia and industry change over time and with technological paradigm shifts?

Existing theories suggest that the closer the knowledge proximity, the more similar the knowledge held by the two parties, and the more conducive this is to knowledge flow (Chen, Mao, Ma, et al., 2024; Dolfsma & van der Eijk, 2016; Sudhindra et al., 2020). In turn, cross-sector knowledge flow may further narrow research gaps (Syafiandini et al., 2024). In this regard, we propose RQ3.

*RQ3:* What is the dynamic relationship between knowledge flow and knowledge proximity? Does this relationship differ across distinct stages of technological development?

The main contributions of this paper are in the following three aspects:

First, we utilize text mining to provide a quantitative analysis method for knowledge proximity between academia and industry. The measurement framework compensates for the lack of traditional research relying on macro indicators to quantify knowledge proximity and provides a reusable quantitative tool for other fields.

Second, we provide evidence of a gradual increase in knowledge proximity between industry and academia. This finding gives micro-level evidence of the symbiotic shift of the two from the asymmetry of institutional logic to the level of knowledge production.

Third, the study confirms that knowledge flow in the field exhibits a reciprocal patter, providing a theoretical basis for optimizing academia-industry collaboration and promoting efficient knowledge flow and sharing.

# 2 Related work

This study focuses on the changes in knowledge proximity between academia and industry in NLP. In this section, we review related work from two perspectives: measurement of knowledge proximity and linkages, and Connections and differences between NLP academia and industry research.

## 2.1 Measurement of knowledge proximity and linkages

Current research on knowledge linkage measurement primarily focuses on citation analysis, semantic mining of text representation, and multidimensional integration based on a complex network.

Citation analysis has its roots in the sociology of science (Price, 1965). The theory of scientific norms proposed by Merton (1973) states that members of an academic community need to recognize and acknowledge the contributions of others by making the basis of their research explicit through citations. In a scientometrics perspective, the process of citation among scientific publications can be abstracted as an explicit representation of knowledge flow (Tsay, 2015). As an ideal proxy indicator of knowledge flow, Lyu et al. (2022) further state that citations between scientific publications or patents represent knowledge flow from the cited entity to the citing entity.

In terms of methodology, the bibliographic coupling approach proposed by Glänzel and Thijs (2011) model scientific literature through a Boolean vector space, where each dimension of the vector represents the citation status (0/1 variable) of papers to specific references. The coupling strength between documents is calculated via cosine similarity, with core literature filtered using specific threshold coupling angles (cos 70° and cos 75°). Glänzel (2012) further introduced the h-index from network degree to reduce the level of subjective intervention. Syafiandini et al. (2024) revealed a complex network structure of knowledge flow in the field of pharmacology through a comprehensive analysis of institutions, institutional types, and biomedical entities, showing that knowledge flow is dominated by academic institutions with bidirectional flow between government and business. Chen et al. (2024) further used dynamic co-citation network analysis, combining community detection algorithms with dynamic network analysis to examine the relationship between internal and external science and technology in enterprises. The results show that the higher the proportion of a company's self-cited papers, the greater its patent output.

In the early stages of text representation, Shibata et al. (2010) pioneered research in the solar cell field, demonstrating that the cosine similarity method with LOGTFIDF vectorization can identify implicit knowledge associations between papers and patents. With the advancement of NLP and AI algorithms, neural network-based modeling encodes complex meanings into dense geometric spaces with minimal distortion, transforming unstructured text into measurable indicators to establish a mapping between theoretical frameworks and empirical data (Aceves & Evans, 2024). An et al. (2021) introduced optimal transport theory to optimize semantic matching, realizing a novel approach for measuring semantic similarity through the theory's core concepts. Alasehir and Acarturk (2022) used the Doc2vec model to quantify the interdisciplinarity of cognitive science and found that its interdisciplinary index increased by 71% over a decade. A subsequent study further constructed an integrated link prediction model to accurately identify potential associations of scientific and technological themes in the pharmaceutical field (Xu et al., 2021). Duede et al. (2024) measured the intellectual distance between articles through Doc2vec text embedding, confirming that universities and research institutions effectively connect people across departments and that cross-departmental collaboration promotes the integration of distant knowledge. Ba et al. (2024) further combined temporal lead-lag distances of themes to detect technological opportunities, proposing four coupling modes between scientific and technological themes and providing a new paradigm for technological opportunity detection.

Complex networks integrate dual information of topological structures and semantic features. Glänzel and Thijs (2017) identify domain core documents by combining literature citations and similarity of noun-phrases extracted by TF-IDF through a joint citation network and text analysis approach. The science-technology bipartite graph constructed by Chen et al. (2023) observed significant overlaps in terminology and themes between patents and papers, indicating existing linkages. Meng et al. (2024) employed K-Core decomposition and network coupling to explore the associations between science and technology in the energy conservation domain. Building on this, Dahlke et al. (2024) used relational networks combined with text embeddings to identify distances between companies and their AI adoption patterns, finding that AI adopters tend to cluster tightly and form relatively closed systems.

Existing studies remain constrained by coarse-grained topic analysis (Aceves & Evans, 2024) and reliance on macro-level indicators like citation frequency, which hinder fine-grained tracking of technological concepts' dynamic evolution. Scholars often depend on proxy metrics such as keyword co-occurrence or are limited by small-sample manual coding. Furthermore, as Glänzel and Thijs (2017) point out, terms and phrases used to express common knowledge bases or general vocabulary inherently lose specificity. These limitations highlight the need to conduct fine-grained entity-level similarity analysis. Parsing micro knowledge units can overcome the ambiguity of traditional macro indicators, offering a more precise analytical dimension to capture the evolutionary trajectories of technical concepts.

For the information and contributions of related work, we have summarized it based on field, data, and methods, as shown in **Table 1**.

**Table 1** Related works on the measurement of knowledge proximity and linkages

| Type | Authors | Field | Data | Methods |
|---|---|---|---|---|
| Citation relationship analysis | Glänzel (2012) | Multiple disciplines in Web of Science | 5,949 papers | Hirsch-type indices |
| | Syafiandini et al. (2024) | Pharmacology | 1,275,536 pharmacology papers. | Citation linkage; PKDE4J; KeyBERT |
| | Chen et al. (2024) | Genetic engineering | 367,901 granted patents and 266,665 papers. | Dynamic network analysis; Louvain |
| Semantic mining of text representation | Shibata et al. (2010) | Solar Cell | 19,063 papers and 10,694 patents published up to 2008 | TFIDF; LOGTFIDF |
| | An et al. (2021) | Thin film head | 1,010 granted patents | SAO; Needleman-Wunsch |
| | Alasehir and Acarturk (2022) | Cognitive science | 118,000 papers from the Annual Meetings conferences | Doc2vec |
| | Xu et al. (2021) | Pharmacology | 9357 papers and 5817 patents | HMM-LDA; CDTM |
| | Duede et al. (2024) | Biochemistry, along with 14 other disciplines | 543,936 papers and 12,149 surveys. | Doc2vec; SEM |
| | Ba et al. (2024) | Energy conservation | 166,643 publications and 91,593 patents were retained | Keybert; Atom Topic Modeling(DATM) |
| Multidimensional integration based on topological structure | Glänzel and Thijs (2017) | Astronomy and Astrophysics | 110,412 papers | Hirsch-type indices TFIDF |
| | Chen et al. (2023) | NLP | 78,292 papers and 171,717 NLP patents | Node2Vec; BERT |
| | Meng et al. (2024) | Energy conservation | 166,643 publications and 91,593 patents were retained. | KeyBert; K-core decomposition |
| | Dahlke et al. (2024) | AI | 1,140,494 company website texts. | BERT; BERTopic |

## 2.2 Connections and differences between NLP academia and industry research

The co-evolution between academia and industry has reshaped technology R&D paths and knowledge production models (Murmann, 2003; Nelson, 1994). It has shattered the unidirectional transmission of knowledge in the traditional linear model (Brescia et al., 2016; Estrada et al., 2016). For instance, the Triple Helix (Etzkowitz & Leydesdorff, 2000; Lawton Smith & Leydesdorff, 2014; Liu & Zhu, 2023), Quadruple Helix (Carayannis & Campbell, 2012), and Knowledge Triangle (Meissner & Shmatko, 2017) all reveal the core feature of bidirectional interaction.

As the fastest-growing subfield in the current AI domain (Haney, 2020), NLP has become a typical scenario of academia-industry co-evolution. The new Transformer paradigm based on the self-attention mechanism proposed by the industry (Vaswani et al., 2017) has reshaped academia's research landscape and spawned applications like the GPT series within four years, demonstrating the strong driving force behind this co-evolution. The comparative analysis of academia-industry in NLP centers on three aspects: paper and patent output, resource allocation, and talent/knowledge flow.

In terms of output characteristics, industry prioritizes technology implementation and commercial value, excelling in patent layout; academia aims to expand knowledge boundaries and advance disciplines, with stronger openness, and mainly produces papers (Gans et al., 2017; Haeussler, 2011). Though academia remains the main source of NLP papers, co-authored works with corporate participation attract more citations (Färber & Tampakis, 2024), and industrial teams hold distinct advantages in impact and technological leadership, dominating the State of the Art (SOTA) models and delivering disruptive innovations (Liang et al., 2024). Abdalla et al. (2023) empirical analysis of 78,187 NLP papers and 701 author resumes confirms that large tech firms' participation drives NLP technological iteration. In contrast, academia focuses on cutting-edge, novel research topics (Chen, Zhang, Zhang, et al., 2024). Regionally, the US and China dominate NLP patent layout, with the US leading in industry and Chinese in academia (Al-Khalifa et al., 2024).

Based on the classic resource dependence theory of Pfeffer and Salancik (1979), organizational strategic choices are constrained by external resource acquisition capabilities. Leading industrial tech giants, leveraging core resources like computing power and data (Ahmed et al., 2023; Färber & Tampakis, 2024), have built strong resource appeal: establishing supercomputing clusters (e.g., Google TPU) with substantial funds, accumulating massive high-quality datasets, and relying on open platforms (e.g., Azure AI, Google Cloud) and open-source ecosystems, they continuously attract knowledge and talent from academic institutions to industry (Shao et al., 2024). This high dependence on industrial resources has further strengthened the industrial orientation of researchers' career choices (Jiang et al., 2024). Stanford AI Index[1] data show that while 40.9% of new AI doctoral graduates entered industry and 41.6% entered academia in 2011, by 2022, the industry proportion surged to 70.7% versus only 20.0% for academia.

Notably, the talent flow is not unidirectionally imbalanced, but rather shows an increasing frequency of bidirectional interaction. By tracking the career trajectories of ACM Distinguished Scientists in computer science (CS), Jiang et al. (2024) identified four core flow paths within and across academia and industry, which have generated significant knowledge spillover effects across departments, regions, and even countries. This mobility is essentially a creative knowledge transformation process: scholars embed tacit research methods and scientific insights into corporate R&D systems (Trippl, 2013), promoting the implementation of technology. At the same time, scholars also adjust their research interests according to new application scenarios (Shao et al., 2024), transforming practical challenges in the industry into academic research issues. This cross-institutional knowledge reorganization enhances the technical reserves of corporate R&D and also fosters more exploratory innovations through the intensive flow of scientific knowledge (Chen, Mao, Ma, et al., 2024; Lian et al., 2025).

[1] https://hai.stanford.edu/ai-index/2024-ai-index-report/education

Despite existing studies revealing academia-industry interaction characteristics from three dimensions. Most focus on case studies of specific regions or institutions and rely on macro indicators, failing to deeply analyze content-level differences between the two. Clarifying the synergy and differentiation between academia and industry from fine-grained knowledge differences is key to understanding the transformation efficiency of academic theories to industrial applications in NLP and identifying their knowledge synergy gaps. Lacking such fine-grained technical measurements to identify gaps, it becomes difficult to optimize the collaboration model, while also risking academia gradually losing its public interest monitoring capability under industry leadership (Ahmed et al., 2023). Based on this, this study combines textual technologies to quantify knowledge proximity from entity and semantic perspectives, aiming to fill existing research gaps.

For the information and contributions of related work, we have summarized it based on field, data, methods, and main findings, as shown in **Table 2**.

**Table 2** Related works on connections and differences between academia and industry.

| Authors | Filed | Data | Methods | Main Findings |
|---|---|---|---|---|
| Liang et al. (2024) | AI | hundreds of thousands of articles from a dataset of AI conferences; 173 notable AI models | Mixed effects regression 、 Excess self-citation analysis | Papers produced by industrial teams are more likely to be disruptive and highly cited, and they propose more SOTA models. |
| Ahmed et al. (2023) | AI | 115234 unique papers、220232 authors | String-matching (Levenshtein) | Industry is increasingly dominating AI research, raising concerns about its control |
| Abdalla et al. (2023) | NLP | 78,187 NLP publications and 701 author resumes | Fuzzy matching | Large tech companies' presence in NLP research has driven the field's development and progress. |
| Färber & Tampakis (2024) | AI | 72,000 pure education papers, 3800 cooperation papers, and 1500 pure company papers | Kruskal-Wallis rank variance test | AI research has indeed become more important than purely academic AI research in recent years. |
| Jiang et al. (2024) | CS | 1444 ACM fellows 267, 804 publications; 25,823 patents related to 205 inventors | Difference-in-difference (DID) | The boundaries between academic research and industrial innovation in the NLP field are blurring. |
| Al-Khalifa et al. (2024) | NLP | 302,934 patents | BERTopic | The United States and China hold dominant positions in the NLP invention landscape, with the US leading in industry and China in academia. |
| Shao et al. (2024) | Various fields | 50,248 scholars and 1,004,464 papers | PrefixSpan; LDA | The research difference between industry and academia has |
| | (includi ng CS) | | | gradually decreases and reached a steady state in recent years. |

# 3 Methodology

The analysis of academia-industry knowledge proximity will be carried out through four dimensions: similarity of entity bag-of-words vectors (explicit overlap of knowledge elements), semantic similarity between text pairs (convergence of semantic conceptual space), coupling of entity networks (network features), and bidirectional flow reinforcement (citation evidence). The research framework specifically includes four core steps, as shown in **Fig. 2**.

Firstly, in the data construction part, we collected the three major conference papers on NLP from 2000 to 2022. By parsing the PDF of the paper and manually annotating it, key fields such as full text, title, abstract, authors, institution information, and year of publication were extracted from the paper.

Subsequently, in the entity and semantic part. As pointed out by Hofstadter and Sander (2013), concepts and their derived conceptual spaces constitute the underlying architecture of human thinking and communication. In addition, knowledge entities serve as the basic unit of knowledge in the literature (Ding et al., 2013). The phenomenon of increased similarity in academia-industry knowledge can be reflected in the overlapping of knowledge elements and intertwining in the conceptual space.

Specifically, we fine-tuned the SciBERT model based on a locally annotated corpus to extract four categories of entities, namely Method, Tool, Metric, and Dataset, with examples presented in **Table A1**. We constructed entity bag-of-words vectors according to year and institution type, and quantified the degree of overlap of knowledge elements using the cosine similarity metric. The complex network approach is further utilized to reveal the dynamic transfer of knowledge dominance in the domain. At the semantic layer, we employed SimCSE unsupervised comparative learning to semantically vectorize text spliced from titles and abstracts. The semantic distance across institution paper pairs is quantified by cosine similarity, the degree of semantic convergence is measured, and technological changes across time stages are tracked.

Finally, for the knowledge flow dimension, based on the bibliometrics perspective, we use citations between papers to measure knowledge flow and reveal knowledge contributions (Lyu et al., 2022). We collected the references of the articles retrieved through OpenAlex[2] and analyzed the trends of citation frequency over time. The dynamic association between changes in knowledge proximity and citations, and the factors that may influence the changes in the distribution of academia-industry citations were explored.

[2] https://openalex.org/

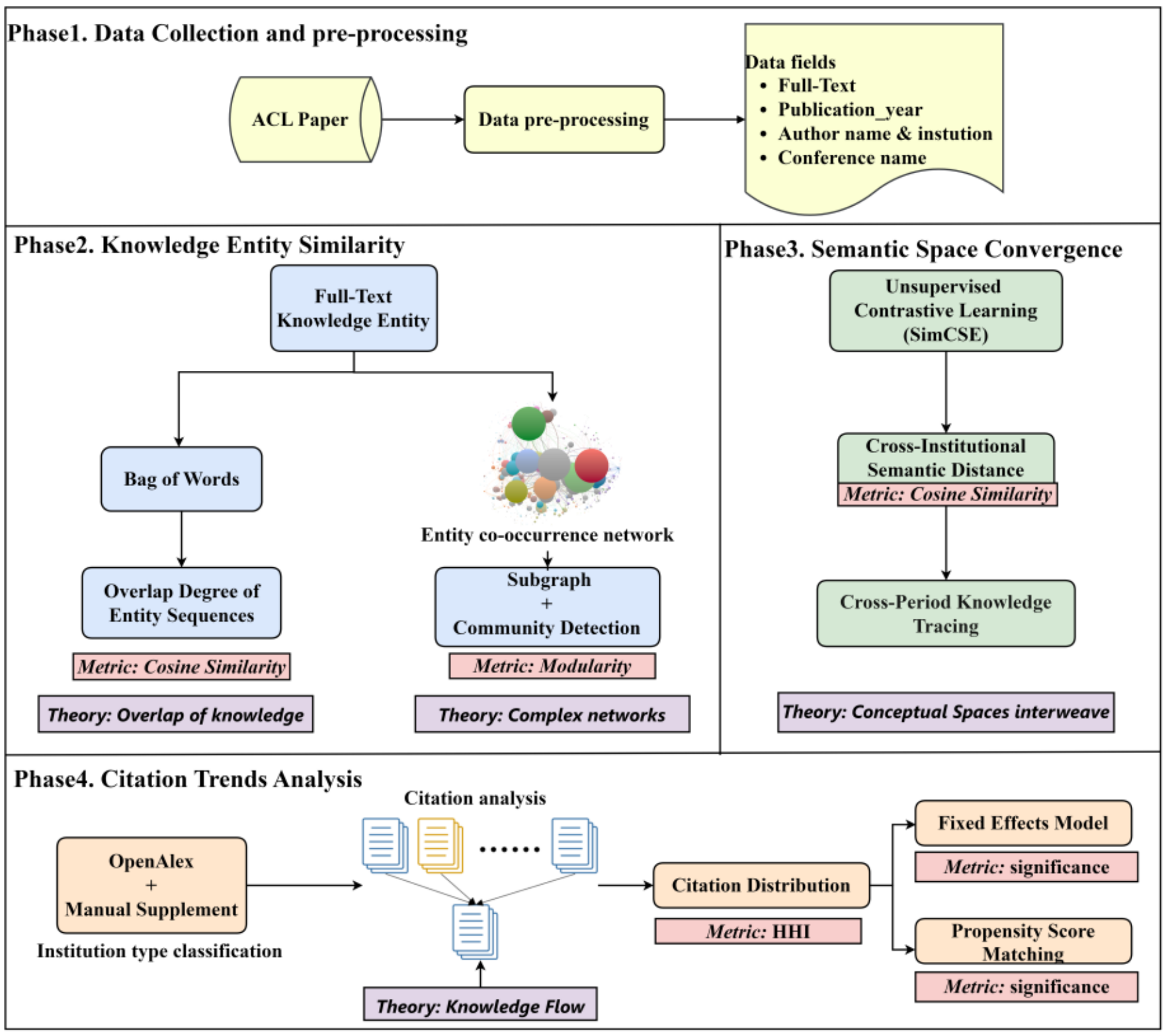

**Fig. 2.** Framework of the knowledge proximity between industry and academia

## 3.1 Field of study

The 2024 Nobel Prize in Physics[3] was awarded to John J. Hopfield and Geoffrey E. Hinton, underscoring the pivotal role of neural networks in computational neuroscience and AI. As one of the fastest-growing segments of current AI, NLP has surged from theoretical exploration to industrial application over the past decade, marked by multiple technological paradigm shifts (Dosi, 1982; Kuhn, 1962). In particular, the launch of ChatGPT by OpenAI on November 30, 2022, marked a breakthrough in NLP technology, pushing NLP to the forefront of technology and the public's vision.

As shown in **Fig. 3**, NLP has evolved from academic foundations to industrial applications over the decades. Hopfield network (Hopfield, 1982) and the BP algorithm (Rumelhart et al., 1986) laid the foundation for modern neural network research. Subsequent iterations led to the emergence of Recurrent Neural Networks (RNN) (Elman, 1990) and Long Short-Term Memory Networks (LSTM) (Hochreiter & Schmidhuber, 1997), which provided a crucial theoretical framework for capturing contextual information in sequential data. The Transformer architecture based on self-attention mechanisms in the industry opened up a new paradigm for NLP tasks, spawning a revolution in the pre-training paradigm (Vaswani et al., 2017).

[3] https://www.nobelprize.org/prizes/physics/2024/summary/

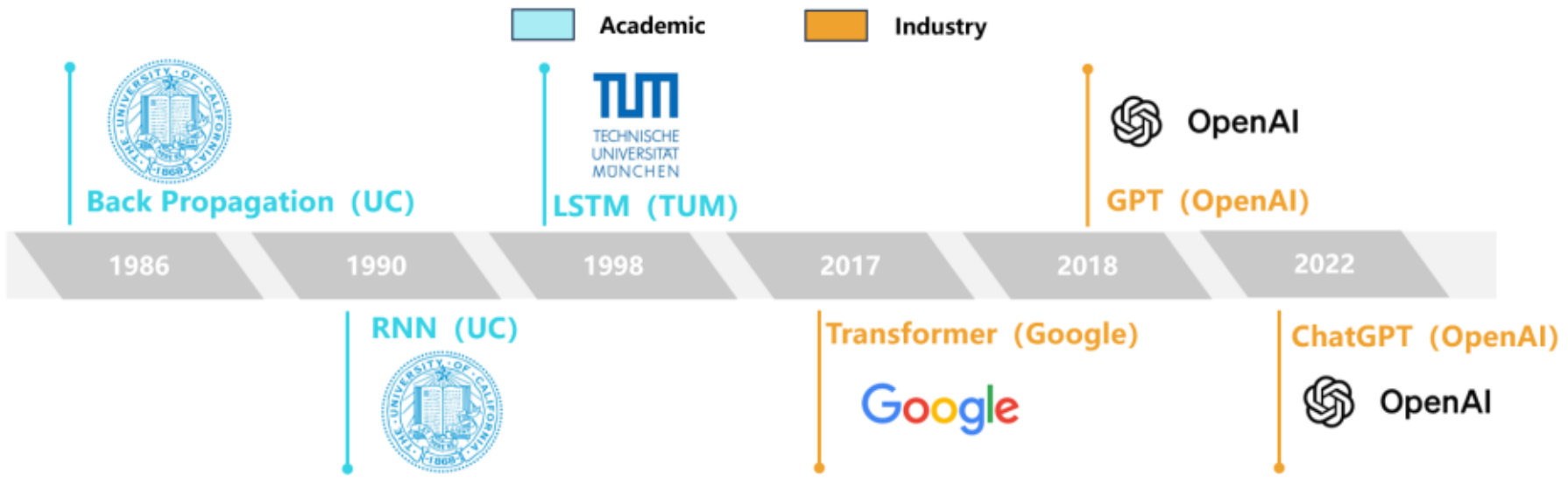

**Fig. 3.** Timeline of ChatGPT Technology Development

The industry and academia in NLP show a triple convergence feature: First, on the strategy of intellectual property rights, while OpenAI consolidates its technological advantages through patents and trade secrets, Meta employs an open source strategy (Apache 2.0 open source protocol) to attract global developers to optimize LLaMa models and drive benchmarking. This open strategy forces companies such as OpenAI to innovate continuously to maintain a voice in the upper reaches of the innovation chain (Larivière et al., 2018), presenting a dynamic game. Second, the mutual mobility of talents (Jiang et al., 2024) fosters cross-domain tacit knowledge transfer, with industry engineering experience reversely driving basic theoretical research. Third, the interpenetration of research paradigms, the pre-training infrastructure built by the industry (e.g., Google's TPUs and Google Cloud) has become a "digital laboratory" for academic research. AI frameworks like TensorFlow and PyTorch significantly reduce the cost of scientific research. Theories such as adversarial training proposed by academics in turn improve the robustness of industrial models.

In view of the above analysis, NLP is a typical field for analyzing the co-evolution between academia and industry. It can enable us to deeply investigate the characteristics of the proximity between academia and industry at the level of fine-grained knowledge.

## 3.2 Data collection and pre-processing

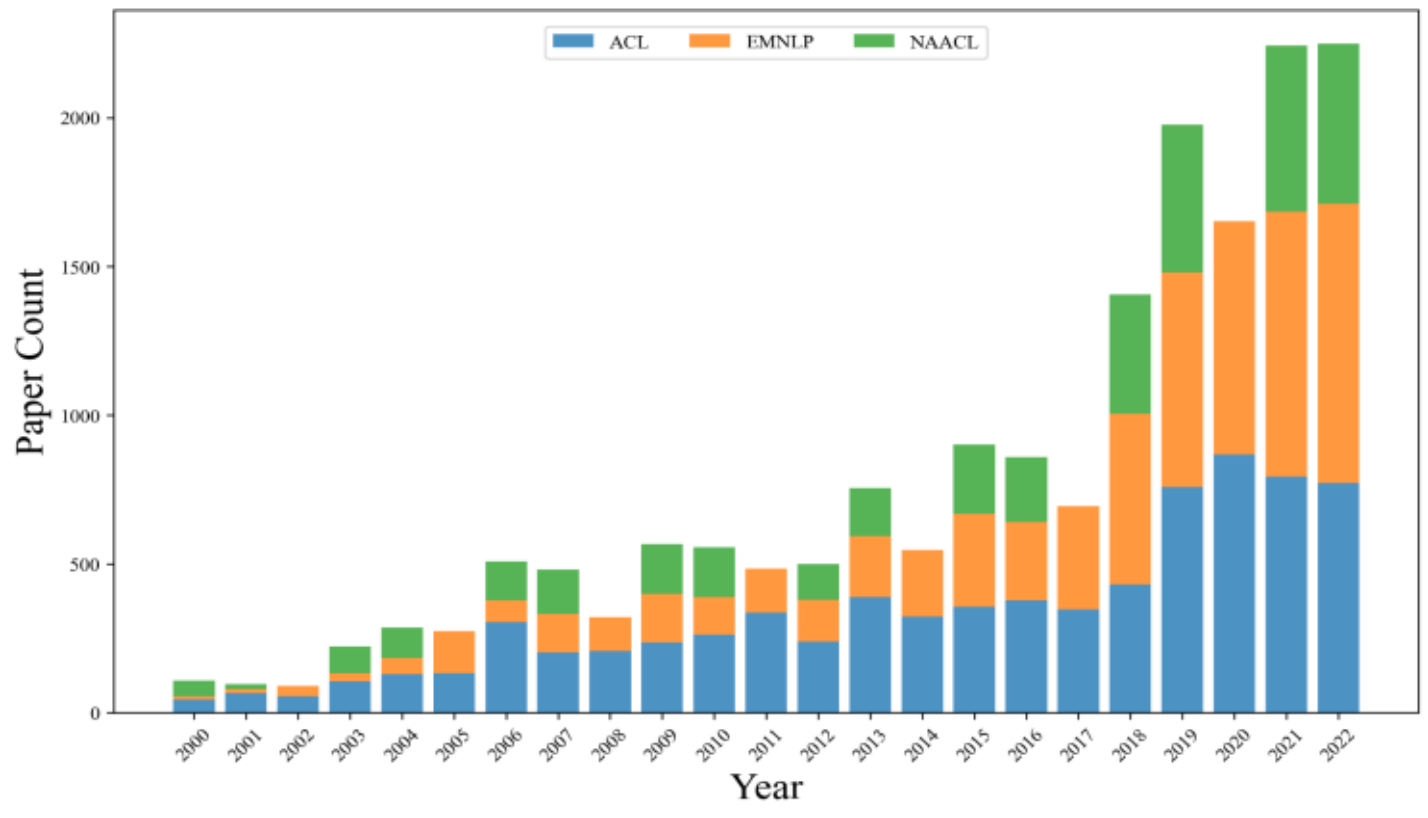

**Fig. 4.** Annual number of papers published at the three conferences.

This study uses full-text papers from the ACL Anthology website[4] as the original dataset, and three major conferences, ACL (Annual Meeting of the Association for Computational Linguistics), EMNLP (Conference on Empirical Methods in Natural Language Processing), and NAACL (North American Chapter of the Association for Computational Linguistics). The full-text data of 17,783 papers were collected from 2000 to 2022. **Fig. 4** shows the number of papers per year for the different conferences, with a general increasing trend. ACL and EMNLP conferences are held annually, while NAACL is not. As the most prestigious conferences in NLP, they not only represent the top level in the field, but also reflect the latest research developments.

### 3.2.1 Entity annotation

NLP is a methodology and data-driven in nature, and most research involves key elements: building or selecting datasets for model training and validation, selecting evaluation metrics to assess model performance and task quality, and utilizing NLP tools (Pramanick et al., 2023; Zhang, Zhang, et al., 2024).

Therefore, we annotated four scientific entities, Method, Tool, Metric, and Dataset, which serve as critical components of the research, clearly reflecting the technological approaches employed in the studies. Among them, the four entities do not overlap with each other. In this study, we randomly selected 50 papers from 17,783 papers and annotated 2,815 sentences containing entities as the annotated corpus, while the remaining papers were used as the full text dataset for extraction. Compared to the SciERC (Luan et al., 2018) and TDM (Hou et al., 2021), containing 2687 and 2010 sentences respectively, the text we annotated is relatively adequate.

The annotation was conducted by five NLP researchers. Under the guidance of a professor in the NLP field. Initially, two annotators pre-annotated the full text of 14 papers. The professor then supervised the resolution of annotation conflicts and the refinement of guidelines. The remaining 36 papers were then annotated by the other three annotators according to these guidelines. To ensure quality, the 36 papers were divided into three groups of 12, and the annotators were paired into three teams ($C_3^2$), with each team responsible for one group. This arrangement ensured that each paper received annotations from two annotators, facilitating consistency checks. The Cohen's Kappa values between the annotators were 0.90, 0.91, and 0.83, reflecting a strong consistency level. Thus, the annotated dataset is considered reliable for training the entity recognition model.

[4] https://aclanthology.org/

### 3.2.2 Technology-related entities recognize and normalize

In this section, we trained the NER model using the locally annotated corpus and employed the best-performing model to infer scientific entities in the remaining papers. We then used hierarchical clustering to normalize the entities, eliminating ambiguities. This normalization provides a more accurate and reliable data source for subsequent experiments.

(1) Training and inference of NER models

With the advancement of Large Language Models (LLMs), their usage in information extraction tasks for scientific texts has surged. However, their performance in specialized domain tasks still lags behind dedicated NER and RE models (Yang et al., 2024). Thus, we continue to utilize pre-trained models for NER tasks. Specifically, we developed a model for identifying scientific entities in papers based on Zhang et al. (2024), leveraging SciBERT (Beltagy et al., 2019)and a cascading binary tagging framework, along with data augmentation techniques.

The model extracts annotated entities from the training set and unannotated abstracts, matching sentences containing these entities as candidate samples. Using the partially trained model, we identify entities in the candidate samples; if all matched entities are recognized, we retain that sentence for data augmentation. This process enhances the diversity of training samples and improves the model's performance.

The specific data augmentation steps are as follows:

We first construct the set of entities $\mathcal{E}$ by setting the original labeled training set to be $\mathcal{D}_{\text{labeled}} = \{(x_i, y_i)\}_{i=1}^{N}$, where $x_i$ is the input text and $y_i$ is the sequence of entity labels; the set of unlabeled abstracts of the paper is $\mathcal{D}_{\text{unlabeled}} = \{s_j\}_{j=1}^{M}$. Extract all entities from $\mathcal{D}_{\text{labeled}}$ to form the set, where $s_j$ is the summary text.

$$\mathcal{E} = \{e_k = (\text{position}_k, \text{type}_k) \mid \forall (x_i, y_i) \in \mathcal{D}_{\text{labeled}}, e_k \in y_i\}_{k=1}^{K} \quad (1)$$

Where $\text{position}_k$ denotes the position of the entity in the text and $\text{type}_k$ is the entity type.

We then perform candidate sample generation by extracting all sentences containing at least one entity $e_k \in \mathcal{E}$ for each unlabeled summary $s_j \in \mathcal{D}_{\text{unlabeled}}$ to generate a candidate set $\mathcal{C}$.

$$\mathcal{C} = \{c_j \subseteq s_j \mid \exists e_k \in \mathcal{E}, \text{position}(e_k) \subseteq c_j\}_{j=1}^{P} \quad (2)$$

Finally, we use pseudo-labeling for verification.

$$\mathcal{C}_{\text{aug}} = c_j \in \mathcal{C} \mid \forall e \in \text{Entities}(c_j), e \in \tilde{y}_j \quad (3)$$

Prediction of candidate samples $c_j \in \mathcal{C}$ using the pre-trained entity recognition model $f_\theta$ to obtain pseudo-labels $\tilde{y}_j$. Samples that satisfy the strict coverage condition are retained, and $\text{Entities}(c_j)$ are all the entities in the $c_j$ that are matched by the $\mathcal{D}_{\text{labeled}}$ matched entities. To obtain the augmented sample $\mathcal{C}_{\text{aug}}$.

In this way, 300 sentences were obtained through data augmentation, and all 3,115 (2,815 + 300) sentences were split into a training set (Trainset), a validation set (Validset), and a testing set (Testset) using an 8:1:1 ratio. Examining the results of model training through Testset ensures the generalization ability of the model without overfitting it.

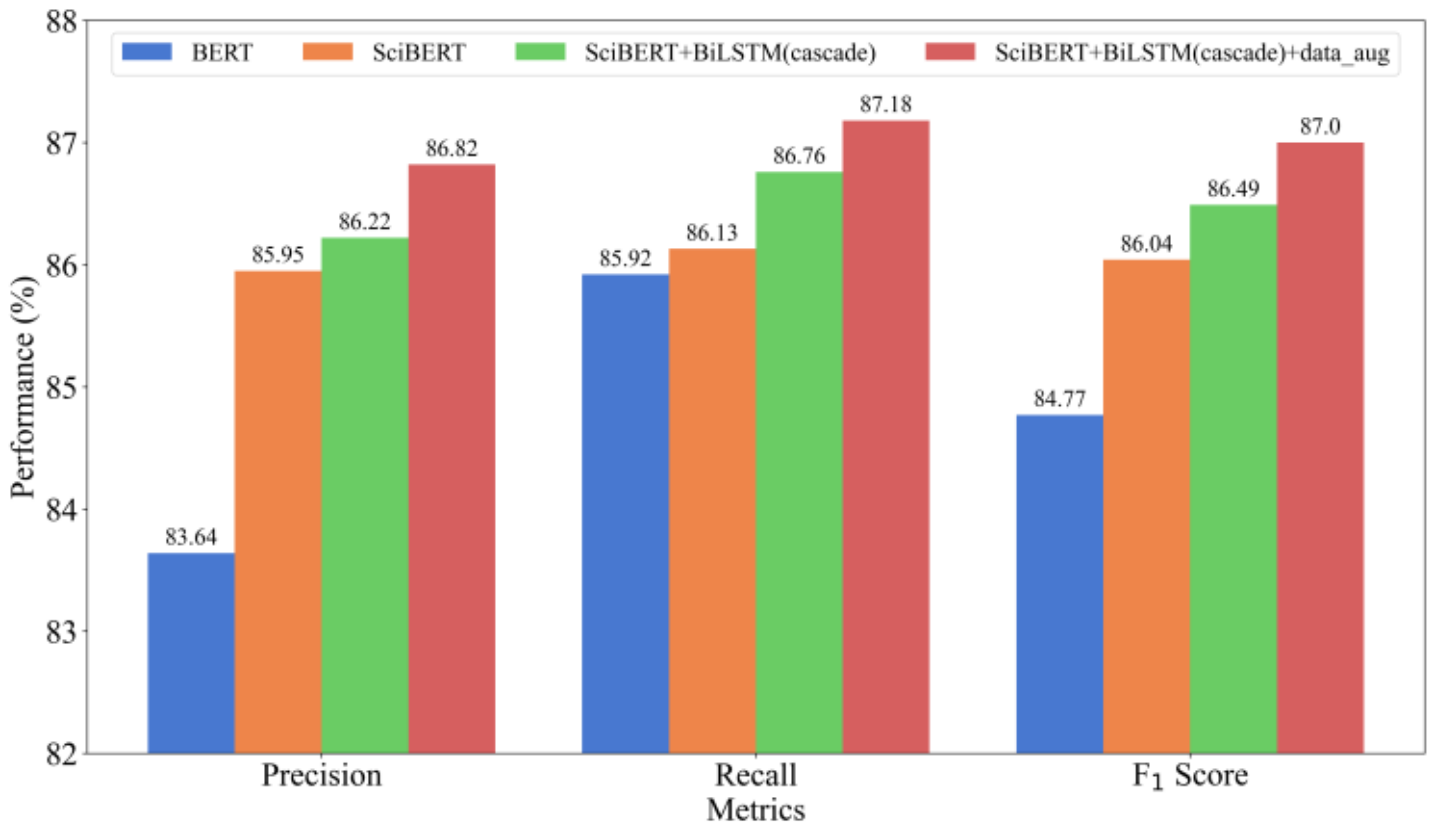

**Fig. 5.** Comparison of different models on our annotated dataset.

**Fig. 5** shows the Precision (P), Recall (R), and $\mathrm{F}_1$ performance of different NER models on the annotated local dataset. The cascading SciBERT+BiLSTM model achieved a $\mathrm{F}_1$ score of 86.49, outperforming the baseline model. With data augmentation, the $\mathrm{F}_1$ score improved to 87.00. We selected the best-performing model for subsequent inference, extracting a total of 534,500 entities from all papers.

(2) Hierarchical clustering-based entity normalization

In the entity normalization process, abbreviations were replaced with full forms, and broad or low-contribution terms, such as “model” and “natural language processing,” were removed. Referring to the method of Zhang et al. (2024), we then performed entity clustering based on edit distance: entities with an edit distance greater than 0.95 were classified into the same cluster, while clusters were merged if their average similarity exceeded 0.8. After normalization, 268,392 entities were obtained. We then filtered out entities with an annual frequency of less than five, considering them as potential misidentifications, and ultimately retained 37,448 valid technology-related entities.

### 3.2.3 Semantic embedding of ACL Anthology papers

To assess the semantic similarity between academic and industrial papers, we excluded papers that did not have abstracts in our dataset of 17,783 papers. The final sample size selected was 17,754 papers, of which 13,225 were academic and industry papers.

For text embedding representation, we employed SimCSE[5] to generate embeddings (Gao et al., 2021). Given that our task involves measuring sentence-level similarity and considering the characteristics of the data (which lacks inherent semantic labels) as well as model performance, we adopted an unsupervised approach. SimCSE improves embedding quality through self-supervised learning and outperforms other models on several Semantic Textual Similarity (STS) datasets.

We perform sentence embedding using contrastive learning by concatenating the title and abstract of each paper into a single text, which is then fed into the model twice. By leveraging the randomness of the dropout layer, we obtain two different embeddings for each sentence. In each batch, the two different outputs of the same sentence are treated as positive examples, while the outputs of other sentences are treated as negative examples. The loss function is:

$$\ell_i = -\log \frac{e^{\frac{\sin\left(\mathbf{h}_i^{z_i}, \mathbf{h}_i^{z_i'}\right)}{\tau}}}{\sum_{j=1}^{N} e^{\frac{\sin\left(\mathbf{h}_i^{z_i}, \mathbf{h}_j^{z_j'}\right)}{\tau}}} \tag{4}$$

$$\mathbf{h}_i^z = f_\theta(x_i, z) \tag{5}$$

Where $f_\theta$ is the encoder function, $z$ is the random mask of dropout, and $\tau$ is the temperature coefficient, which is used as a temperature coefficient to regulate the shape of the probability distribution in the vector space, and $\tau$ is set to be 0.01 in this experiment. A lower $\tau$ makes the semantic vector probability distribution of the text distributed in the space more steeply, and makes the model to increase the distance between positive and negative samples (Gao et al., 2021; Wang & Liu, 2021). In the unsupervised task scenario of this study, a strict distribution of positive and negative samples helps to provide more discriminative embedding vectors for the subsequent semantic similarity computation task.

We embedded paper titles and abstracts using our fine-tuned sup-simcse-bert[6] model.

### 3.2.4 Division of types of publishing institutions

Chen et al. (2024) determined the author's institutional type by manually annotating institutional information and integrating the Global Research Identification Database(GRID)[7] database. Based on the institutional information they had already obtained, this study further supplemented the institutional data of the papers by combining manual annotation with OpenAlex queries for the entries in the dataset used in the research that had not yet obtained institutional information. It should be noted that OpenAlex institutional classification criteria are consistent with the GRID classification standards, still based on the Research Organization Registry (ROR)[8] system. After GRID became an integrated part of Dimensions, ROR became the continuation of GRID. The system assigns a unique organization identifier (ROR ID), to each institution. Ultimately, we completed the author and corresponding institutional information for 17,783 papers.

[5] https://github.com/princeton-nlp/SimCSE

[6] https://huggingface.co/princeton-nlp/sup-simcse-bert-base-uncased

[7] The Global Research Identification Database (GRID) classifies institution types into eight categories: government, education, company, facility, healthcare, nonprofit, archive and other.

**Table 3** Distribution of types of institutions

| Institution_Type | Frequency | Ratio |
|---|---|---|
| Education | 49672 | 69.91% |
| Company | 14063 | 19.79% |
| Facility | 3776 | 5.31% |
| Nonprofit | 1966 | 2.77% |
| Government | 902 | 1.27% |
| Healthcare | 320 | 0.45% |
| Other | 296 | 0.42% |
| Archive | 51 | 0.07% |
| Total | 71,046 | 100.0% |

Given the research objective of this paper, which is the proximity of knowledge between academia and industry, we focus our attention on academic and industrial institutions. First, we classify education and company institutions into academia and industry, respectively. Following the definitions from existing studies (Chen, Zhang, Zhang, et al., 2024; Xu et al., 2022), healthcare institutions primarily include research-oriented medical institutions and government-funded research centers (e.g., Harvard Medical School, Mayo Clinic). Thus, we categorized education and healthcare into the academia category and company into the industry category. The final distribution of institution types is shown in **Table 3**, where academic and industry institutions together comprise 90.15% of the dataset, while other types of institutions are excluded (less than 10% of the total). In addition, when dealing with the complexity of having multiple institutions as authors, we follow Hottenrott et al. (2021) and consider the first institution listed by the authors as their primary affiliation.

For paper type classification, a paper is classified as: (a) "Academic" if all authors are affiliated with academic institutions; (b) "Industry" if all authors are affiliated with industry institutions; (c) "Academic-

[8] https://ror.org/

Industry cooperation" (hereinafter referred to as cooperation papers) if both academic and industry institutions are represented among its authors.

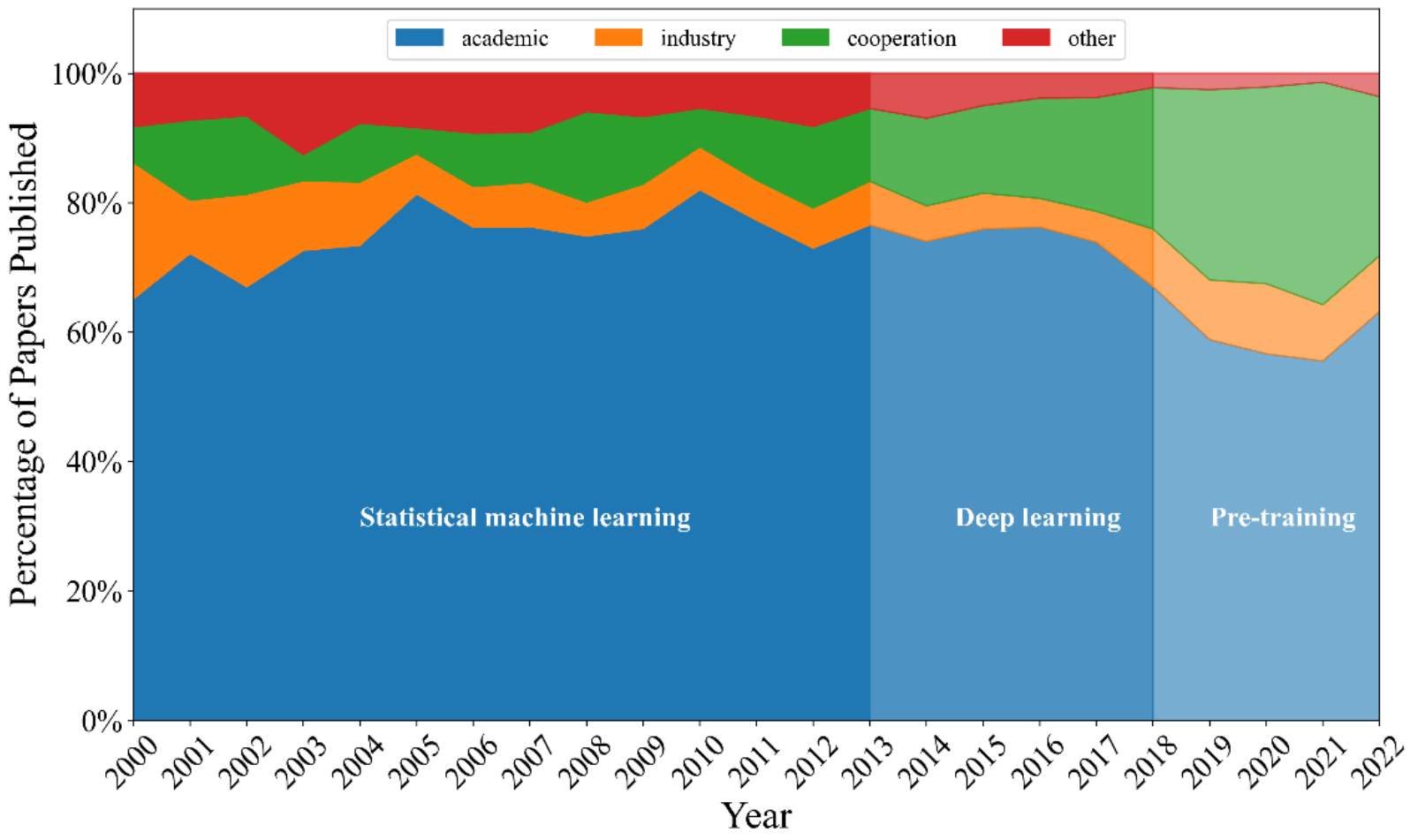

**Fig. 6.** Distribution of publications by different types of institutions

**Fig. 6** illustrates the annual publication proportions of different institutions for the 17,783 papers. Additionally, we divided 2000–2022 into three periods following NLP field development (Zhou et al., 2020).

## 3.3 Knowledge proximity measurement based on knowledge entities and semantic space

In this section, we build fine-grained knowledge proximity metrics to address RQ1

### 3.3.1 Knowledge entity similarity analysis

Utilizing the entities identified in Section 3.2.2, we used a bag-of-words model to represent these entities as two separate sequences for each year. After alignment, we generated the corresponding bag-of-words vectors. By calculating the cosine similarity between these vectors, we assessed the similarity of entities across different years, resulting in a similarity matrix.

The similarity matrix provides an intuitive way to assess the knowledge proximity between industrial and academic research in a given year. This approach accounts not only for the presence of different entities but also incorporates their frequencies.

For example, when calculating the similarity between industrial research in year1 and academic research in year2 (where year1 and year2 may be the same), we define the vocabulary of the bag-of-words model using the union of the entities $|\boldsymbol{V}|$. $N$ represents the total number of entities from both industrial and academic sectors, which is the size of the union $|\boldsymbol{V}|$.

$$\boldsymbol{Bow_{iy}} = \langle F_{iy}(e_0), F_{iy}(e_1), \ldots, F_{iy}(e_{N-1}) \rangle, e_0, e_1, \ldots, e_{N-1} \in |V_i|, i \in \{0,1\} \quad (6)$$

Here, $Bow_{iy}$ represents the bag-of-words vector for either the industrial or academic sector in year $y$, where $i = 0$ denotes the industrial sector and $i = 1$ denotes the academic sector. $F_{iy}$ represents the frequency of the entity's occurrence in year $y$ for sector $i$.

$$F_{iy}(e_j) = \begin{cases} q, & e_j \in |V_i| \\ 0, & otherwise \end{cases} \tag{7}$$

Cosine similarity was used to calculate the similarity of entity series in academia and industry in different years.

$$\text{Sim}(y_1, y_2) = Cos(\boldsymbol{Bow_{0y1}} \cdot \boldsymbol{Bow_{iy2}}) = \frac{\boldsymbol{Bow_{0y1}} \cdot \boldsymbol{Bow_{iy2}}}{\|\boldsymbol{Bow_{0y1}}\|_2 \cdot \|\boldsymbol{Bow_{iy2}}\|_2} = \sum_{k=0}^{N-1} \frac{F_{0y_1}(e_k)F_{1y_2}(e_k)}{Z_{0y_1}Z_{1y_2}} \tag{8}$$

where $Z_{iy} = \sqrt{\sum_{k=0}^{N-1} F_{iy}(e_k)^2}$, and $N$ represents the dimension of the vector.

### 3.3.2 Convergence of semantic space

Using the contrastive learning method from Section 3.2.3, we generated a vector representation for each paper and calculated the cosine similarity between all pairs of academic and industrial papers.

$$s(p_i, p_j) = \frac{h_{z_i} \cdot h_{z_j}}{\| h_{z_i} \| \| h_{z_j} \|} \tag{9}$$

$s(p_i, p_j) = \frac{h_{z_i} \cdot h_{z_j}}{\|h_{z_i}\|\|h_{z_j}\|}$, where $h_{z_i} = f_\theta(x_i, z), h_{z_j} = f_\theta(x_j, z)$. The similarity between the vectors is calculated by cosine similarity, where $p_i, p_j$ are the two papers to be compared, $z$ is the random dropout, $x_i$ is the text, $f_\theta$ is the model encoder.

In calculating the similarity between different years, we adopted the concept of nearest neighbors. However, considering the inconsistency in publication volumes across different years, this study utilizes a threshold as a substitute. Specifically, we measure the threshold by comparing each paper with its counterpart from the opposing institution.

$$s(p_i, p_j) = Q_{0.9}{}^{(y_1, y_2, L)} = Quantile(\{ s(p_i, p_j) \mid p_i \in \mathcal{D}_{y_1}{}^{L}, p_j \in \mathcal{D}_{y_2}{}^{1-L} \}, 0.9) \tag{10}$$

Where, $\mathcal{D}_y{}^L$ represents the set of all papers of institution type $L$ of institution type in year $y$, $s$ represents the semantic vector cosine similarity between the focal paper $p_i$ and the paper $p_j$ of the opposing institution.

In this formula, $L = 0$ represents the industrial sector, while $L = 1\ represents$ the academic sector. Subsequently, we compute a set of similarity values by comparing each paper with all publications from the opposing sector for the target year and compare the 90th percentile (top 10%) of these values against the threshold, which is referenced from existing literature (An et al., 2021; Jeon et al., 2022; Liu et al., 2024;

Mukherjee et al., 2017). Then, we calculated the overall paper similarities and analyzed their distribution (**Fig. 7**). Based on this distribution, the 90th percentile of the overall similarity score was 0.736, which was set as the high similarity threshold.

$$P(y_1, y_2, L) = \frac{\left|\left\{ s(p_i, p_j) \mid p_i \in \mathcal{D}_{y_1}{}^{L}, p_j \in \mathcal{D}_{y_2}{}^{1-L}, s(p_i, p_j) > 0.736 \right\}\right|}{\left|\mathcal{D}_{y_1}{}^{L}\right|} \quad (11)$$

We regard the proportion $P$ of high similarity as the similarity between industry and academia in different years. The numerator represents the number of paper pairs that satisfy $s(p_i, p_j) > 0.736$. The denominator $\left|\mathcal{D}_{y_1}{}^{L}\right|$ is number of institution papers.

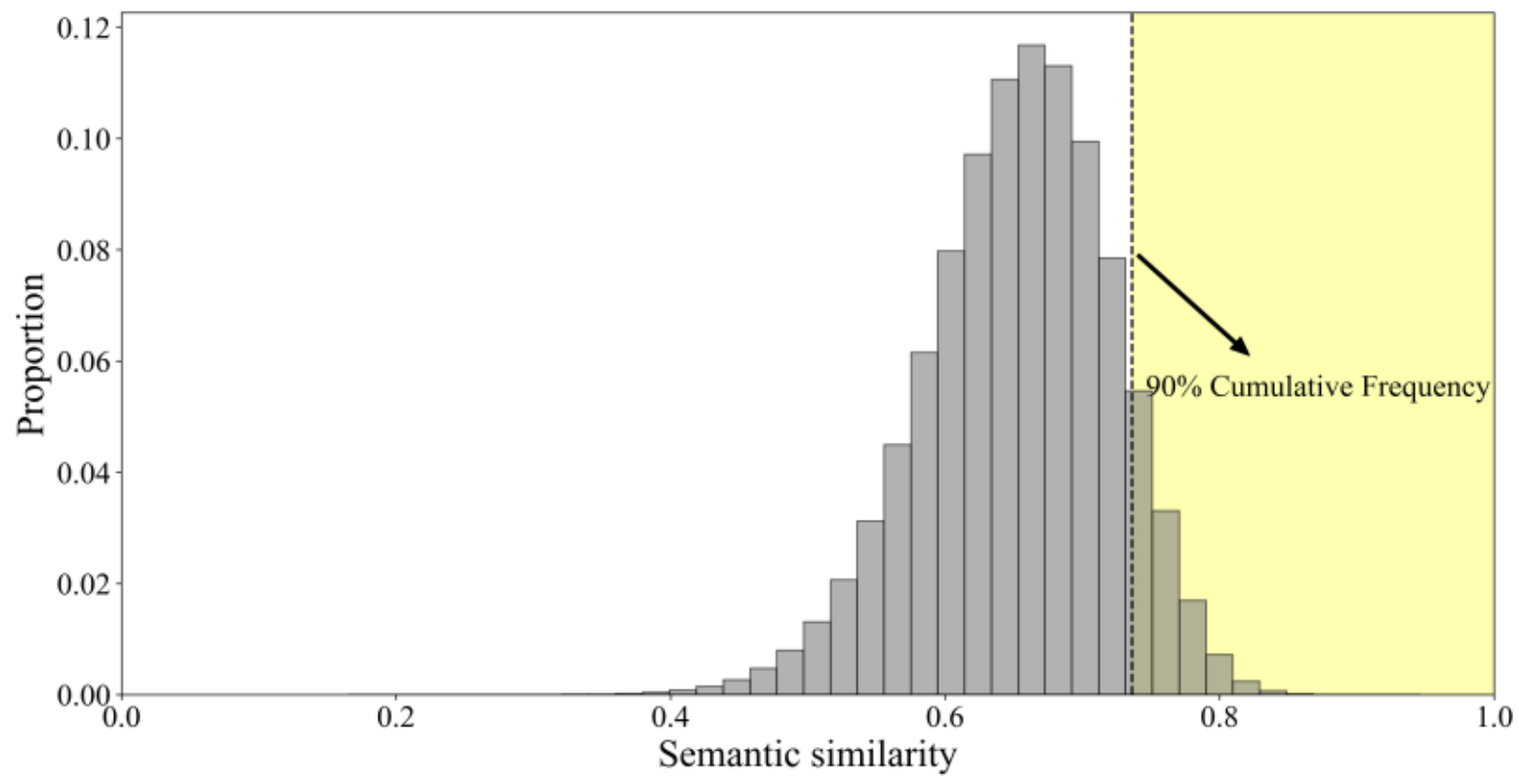

**Fig. 7.** Probability distribution of cosine similarity for papers in the dataset

Specifically, using 2018 industrial papers and 2019 academic papers as an example, we first calculate the similarity between each 2018 industrial paper and every 2019 academic paper. For each 2018 paper, we construct a set of similarity values and extract the 90th percentile as its similarity score with 2019. Based on this, we set 0.736 as the high similarity threshold and compute the proportion of 2018 industrial papers with a similarity score above 0.736. This proportion quantifies the similarity of 2018 industrial research to 2019 academic research.

### 3.3.3 Knowledge entity co-occurrence network analysis

According to Barabasi and Albert's (1999) complex network theory, highly influential nodes attract more new nodes. With the continuous evolution and transformation of technological paradigms, new entities continue to emerge in the NLP field (Zhang, Zhang, et al., 2024). During the process where paradigm shifts drive the reconfiguration and renewal of knowledge systems, such changes are intuitively reflected in the topological characteristics of the network.

Using each year as a time unit, we treated entities mentioned in papers as connected to build an entity co-occurrence network. We calculated the largest connected component (LCC) for different institutions, representing highly related entities and their relationships (Luo et al., 2023). Analyzing the LCC reveals key components and major linkage paths in the knowledge network structure. We constructed the connected component of the entire field based on all entities. Furthermore, we separately removed publications from three types of institutions to study the missing scale of the giant connected component, to measure the control strength and influence of different types of institutions on the overall knowledge network structure, and to analyze the sensitivity of the network structure.

Further, this study focuses on the common nodes in the LCC of the three types of institutions, analyzed in terms of the technology portfolio patterns as well as the contribution to knowledge connectivity:

$$V_{common} = V_A \cap V_I \cap V_C \tag{12}$$

where $V_A, V_I, V_C$ represent the set of nodes in the subgraphs of academia, industry and cooperation, and $V_{common}$ is the common node between the three.

$$E'_T = (u, v) \mid (u, v) \in E_T \ and \ u \in V_{common} \ and \ v \in V_{common} \tag{13}$$

$E'_T$ represents the set of concatenated edges that exist in the public node for the type $T$.

Additionally, this study was based on the Louvain algorithm (Blondel et al., 2008) for modularization calculations and community structure analysis. By calculating the network's modularity (Modularity), this indicator quantifies the difference between the connection density within communities and the sparsity of connections between communities, effectively assessing the cohesiveness and separation of community structures in the network.

$$Q = \frac{1}{2m} \sum_{i,j} \left[ W_{ij} - \frac{k_i k_j}{2m} \right] \delta(c_i, c_j) \tag{14}$$

Where $Q$ is the modularity, $m$ denotes the number of edges in the network, $\delta(c_i, c_j)$ takes 1 when $i$ and $j$ belong to the same community (and 0 otherwise), $k_i$ represents the degree of node $i$, $W_{ij}$ is the adjacency matrix of the network. High modularity indicates strong intra-community connectivity and barriers between communities, while a low modularity degree reflects strong knowledge circulation between communities.

## 3.4 Citation trends analysis

### 3.4.1 Retrieval of references and their institutional information

The process of citation between scientific publications can be abstracted as an explicit characterization of knowledge flow (Tsay, 2015). Measuring citations between publications is a common approach to knowledge flow (Lyu et al., 2022). At the knowledge flow level, we queried references in 17,783 papers through OpenAlex and typed the affiliations of authors.

In the case of missing institution types in OpenAlex (Zhang, Cao, et al., 2024), i.e., for some of the queries, OpenAlex returns null institutions (Completely missing institutional information). We use the author's ID to run a subsequent query and retrieve all institutional affiliations, matching by publication year and the author's institutional tenure. Through the above steps, we added some papers with missing institutions. This is consistent with the strategy of (Zhang, Cao, et al., 2024) in missing information processing.

In the end, we obtained a total of 479,923 references. And for those cited papers that appeared multiple times in our data, we supplemented their institution information type by searching again manually. In the end, we obtained 466,856 papers containing institution information.

By counting the frequency of occurrence of these eight categories (including multiple occurrences of the same institution), the statistical results are shown in **Table 4**. Among the 8 categories, "education" accounts for the largest proportion at 67%. Consistent with the research category classification criteria described earlier (Chen, Zhang, Zhang, et al., 2024; Xu et al., 2022), we classified "education" and "healthcare" as academic institutions, and "company" as industrial institutions. The cumulative frequency of "education", "healthcare", and "Company" reached 87.29%, while the remaining categories had lower proportions and were collectively excluded.

The criteria for determining paper types also align with previous definitions: papers where all authors are affiliated with academic institutions are classified as "academic papers"; those with all authors from industrial institutions are "industrial papers"; and papers with authors from both academia and industry are cooperative.

**Table 4** Cumulative frequency proportions of different institutions in the references

| Institution_Type | Frequency | Ratio |
|---|---|---|
| education | 1137566 | 67.05% |
| company | 309657 | 18.25% |
| facility | 112446 | 6.63% |
| nonprofit | 55554 | 3.27% |
| government | 34437 | 2.03% |
| healthcare | 33687 | 1.99% |
| other | 11373 | 0.67% |
| archive | 1948 | 0.11% |
| Total | 1696668 | 100.0% |

### 3.4.2 Analysis of citation distribution characteristics

Further, we measure the citation characteristics of both academia and industry in the field of NLP, and we explore the tendency of industry and academia to cross-cite and self-cite using the excess self-citations (ECC) model, with specific reference to (Liang et al., 2024).

$$ECC_t(L_i) = \left[P\left(L_i \mid L_i, t \leq t_p\right) - P\left(L_i \mid t \leq t_p\right)\right] - \left[P\left(L_j \mid L_i, t \leq t_p\right) - P\left(L_j \mid t \leq t_p\right)\right] \tag{15}$$

Where $L = 0$ represents industry, and $L = 1$ represents academia, $ECC_t(L_i)$ denotes the degree of excessive self-citation in papers published by institution type $i$. $L_i$ and $L_j$ refer to different institutions, meaning $L_i = L_j \oplus 1$. $t_p$ represents the publication time of the paper, and $P(\alpha \mid \beta, t \leq t_p)$ indicates the probability of institution type $\beta$ citing papers from institution type $\alpha$. $P(\alpha \mid t \leq t_p)$ represents the probability of any paper citing those from institution type $\alpha$, without specifying the institution type.

We also measured the citation distribution uniformity for each year to understand citation exchanges between academia and industry. Specifically, we used the Herfindahl-Hirschman Index (HHI) to assess the concentration of citation distribution. The HHI is a standard metric for measuring concentration.

$$\text{HHI} = \sum_{i=0}^{2} \left( \frac{citation_count(L_i)}{citation_count(L_{total})} \right)^2 \tag{16}$$

Where $L = 2$ represents collaboration. $Citation_count(L_i)$ refers to the number of citations from institution type $i$. A higher HHI value indicates greater concentration, while a lower value suggests more balanced citation contributions, reflecting bidirectional knowledge flow between academia and industry.

When we analyzed the correlation between citation distribution and knowledge proximity. We first conducted an overall analysis by calculating the annual mean HHI index of paper citation distribution and comparing it with the corresponding annual entity similarity.

Subsequently, at the semantic level, we utilized a multiple regression model based on a single paper as the unit of analysis. The semantic similarity between each paper and its opposing institution type is set as the dependent variable, and the HHI of its citation distribution is set as the independent variable. Specifically, for a paper $p_i$ published in year $y$, its semantic similarity is defined as the top 90th percentile of semantic similarity to $p_j \in \mathcal{D}_y{}^{1-L}$ in its opposing institution. The calculation method is consistent with Eq. (13), limiting the analysis to papers of the same year only, as follows:

$$\text{SemSim}_i = Quantile\left(\left\{ s(p_i, p_j) \mid p_i \in \mathcal{D}_y{}^{L}, p_j \in \mathcal{D}_y{}^{1-L} \right\}, 0.9\right) \tag{17}$$

Regarding the setting of control variables, research team-level characteristics have an important impact. Large R&D teams tend to have a broader and richer knowledge base, and collaboration among team members can increase opportunities for cross-domain knowledge integration (Wu et al., 2025). At the same time, such teams will focus on current hot areas and cutting-edge results on which to conduct development research (Wu et al., 2019), so we include the number of authors and institutions as control variables. In addition, we measure the flow of knowledge using the citation distribution feature, which affects the calculation of the indicator HHI, and we also consider the number of citations as a control variable. Also referring to (Wu et al., 2025), we applied a logarithmic transformation for continuous control variables to deal with their skewed distributions.

Finally, taking into account the differences between disclosure mechanisms and institutions (Gans et al., 2017), we also control for the fixed effects of institution type to distinguish between institutional characteristics of different subjects. The specific regression model is as follows:

$$\mathrm{SemSim}_i = \alpha + \beta_1 \mathrm{HHI}_i + \beta_2 \ln(\mathrm{InstNum})_i + \beta_3 ln(\mathrm{AuthNum})_i + \beta_4 \ln(\mathrm{Ref})_i + \sum_{k=0}^{1} \delta_k \cdot Instype_k + \sum_{\tau=2000}^{2022} \gamma_\tau \cdot D_\tau + \varepsilon_i \quad (18)$$

Where $D_\tau$ is a year dummy variable to absorb time trends. $i$ denotes a single paper and $\varepsilon_{it}$ is a randomized perturbation term. $Instype_k$ represents the institution type, with $k = 0{,}1$ corresponding to industry and academia, respectively.

# 4 Results

In this section, we answer the three research questions mentioned in the introduction, based on the results obtained from text mining and citation analysis in Chapter 3.

## 4.1 Research content proximity between industry and academia

From the two analytical dimensions of entity similarity and semantic similarity, this section examines the changing characteristics of research content similarity between industry and academia, thereby addressing research question RQ2.

### 4.1.1 Changes in knowledge entities' similarity within ACL Anthology papers

In this article, scientific entities were extracted from the full text and categorized by year and institution. Based on this, bag-of-words vectors were constructed as described in equation (11), and the cosine similarity between these vectors was calculated to quantify the proximity between academia and industry.

By measuring the cosine similarity of bag-of-words vectors for entities from 2000 to 2022, we generated the heatmap. **Fig. 8** (a) displays results based on all entities, while **Fig. 8** (b) focuses exclusively on method entities. Isolating method entities makes sense because they specifically represent the research approaches, techniques, and models employed.

We observe a generally gradual upward trend in entity similarity between industry and academia. Before 2014, the similarity formed a near-rectangular pattern, indicating stable technological trends and a high degree of alignment between the two sectors, reflecting the relative stability of the technological paradigm during this period. Despite the initial development of deep learning technology, it has not yet had a significant impact on the established research paradigm, and progressive innovation in the technological field was still constrained by the original research paradigm.

Since 2018, the trajectory of the heat map **Fig. 8** shows a clear shift, one of which is the rapid rise in similarity, breaking the smooth growth pattern of the previous period, with the level of similarity improvement significantly surpassing the previous period. Secondly, the similarity maintenance cycle is significantly narrowed, reflecting a clear difference with the traditional technology evolutionary rhythm, and research content spaced only three years apart shows significant similarity decay. This coincides with the period of the pre-training paradigm brought about by the emergence of the Transformer architecture, BERT and GPT, among others (Devlin et al., 2019; Radford & Narasimhan, 2018; Vaswani et al., 2017), reflecting the discontinuous nature of the scientific revolution in this paradigm (Dosi, 1982; Kuhn, 1962).

Furthermore, we analyzed the dynamic differences in knowledge proximity between academia and industry from the perspectives of three entity types: Tools, Datasets, and Metrics. The results are shown in **Fig. B1**. Tools and Datasets exhibit stage-driven similarity growth patterns. Tool entities' similarity increases highly synchronized with technological paradigm shifts. Datasets show high similarity in consecutive years but exhibit noticeable fluctuations across different years, reflecting the dynamic adaptation between dataset iterations and technological demands. The sustained high similarity among metric entities stems from standardization. A unified evaluation framework provides directly comparable and verifiable benchmarks, fostering institutional research collaboration while ensuring alignment between industry and academia in technological evolution (Aufrant, 2022).

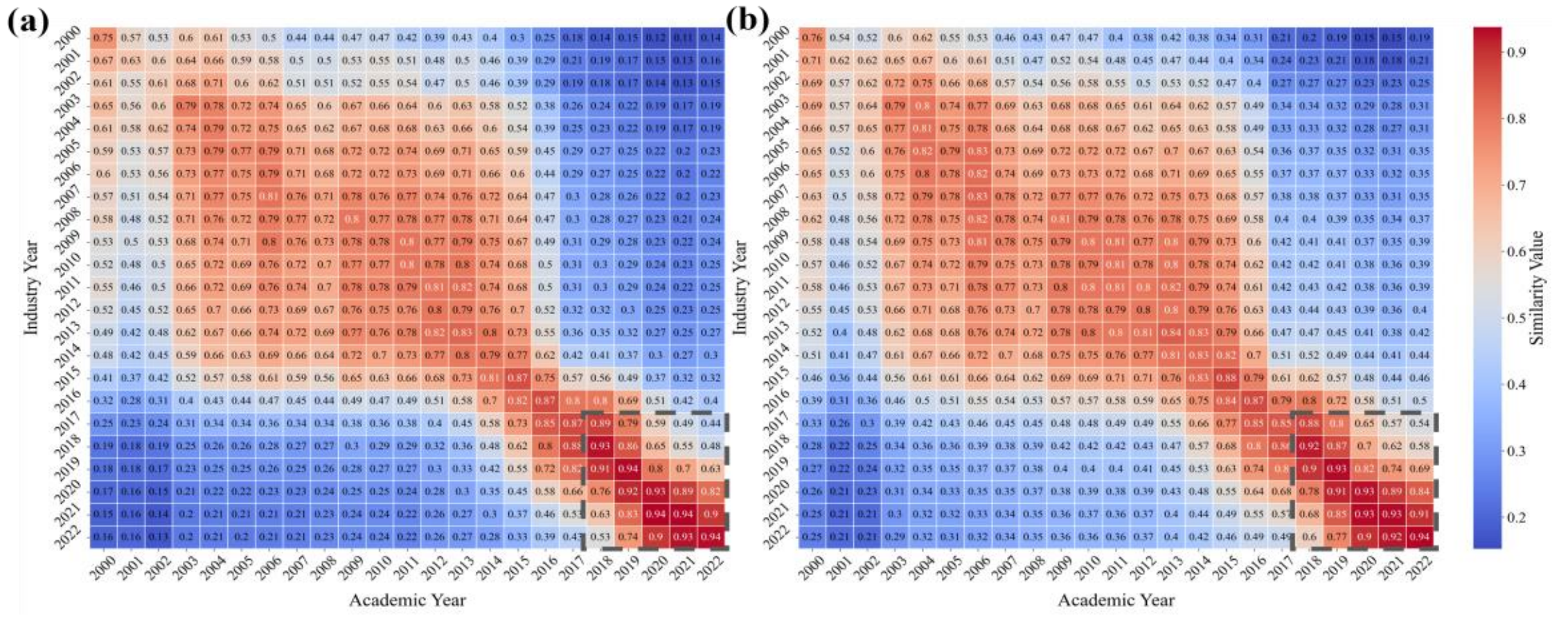

**Fig. 8.** Heatmap of bag-of-words vector similarity for entities. (a) The heatmap of bag-of-words vector similarity is based on all entities; (b) The heatmap of bag-of-words vector similarity is based on the method entities.

*Notes.* The black dotted box in the lower right corner represents the pre-training paradigm stage after 2018, which is the brightest area in the heat map

Further, we analyzed traditional macro-level indicators of academia-industry proximity, as measured by the frequency of collaboration in publications (Blankenberg & Buenstorf, 2016; Fischer et al., 2019). The trend of similarity based on the level of the knowledge entity is consistent with the percentage of collaborative papers, as reflected in **Fig. 9**. Both in terms of collaboration frequency and knowledge proximity gradually increase over time, reaching the highest level in the entire observation cycle during the pre-training phase.

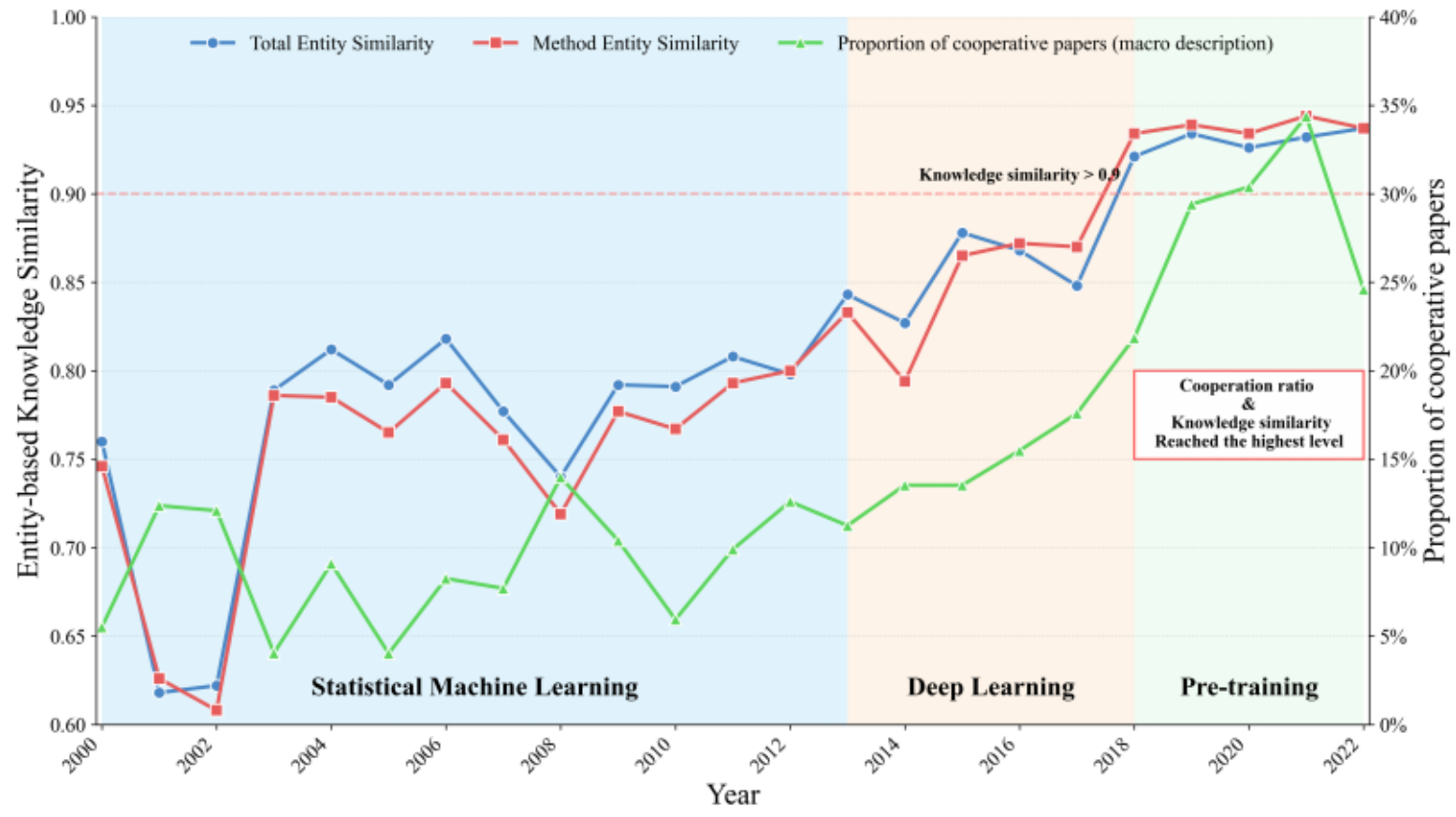

**Fig. 9.** Trends in entity knowledge proximity and proportion of collaborative papers (2000 - 2022)

To further verify the reliability and stability of the similarity index based on knowledge entities, the Coefficient of Variation (CV) test was used (see **Fig. C1**). A smaller CV means the index has smaller fluctuations in measurement results under different sampling conditions, effectively avoiding biases from random sampling. Compared with traditional macro indicators, the indicators based on knowledge entities maintain a smaller CV in different sampling scenarios. This not only confirms the greater stability and reliability of fine-grained measurement method, but also verifies its significant stability advantage over traditional macro-indicators.

### 4.1.2 Semantic proximity and new opportunities brought by technological innovations

Building on the overall perspective of entity measurements between institutions, we further calculated the similarity between each academic or industrial paper and papers from the counterpart institutional type at the individual paper level. By setting a threshold, we computed the proportion of highly similar papers to describe the overall similarity between the two types of institutions.

Compared to entity-based sequence comparisons, semantic analysis takes context and multi-level semantic relationships into account, making the results of semantic similarity calculations generally more complex. **Fig. 10** (a) and **Fig. 10** (b) show a similar trend with entity similarity, with higher similarity values in the lower right region. To ensure the robustness of the results, we also supplemented the experiments at the 5% and 20% thresholds (see **Fig. B2**).

Especially when calculating the similarity of industry papers relative to academia, the similarity value is more pronounced than the similarity of academic papers to industry. This may be attributed to the larger number of academic papers, resulting in higher similarity when industry searches for academic articles. Additionally, the broader research directions in academia allow industry papers from various fields to be more easily comparable.

Based on the vertical analysis of **Fig. 10** (a) and the horizontal analysis of **Fig. 10** (b), we observe significant highlighted areas in academia around 2009 and 2014. A similar phenomenon is noted around 2018; however, the impact of frequent technological iterations results in less pronounced similarity compared to 2009 and 2014. These studies demonstrate a high degree of similarity over an extended time frame, with research content building on previous work, particularly in close relation to studies from earlier years. Within the context of NLP development, these years also represent critical technological turning points and differentiation phases.

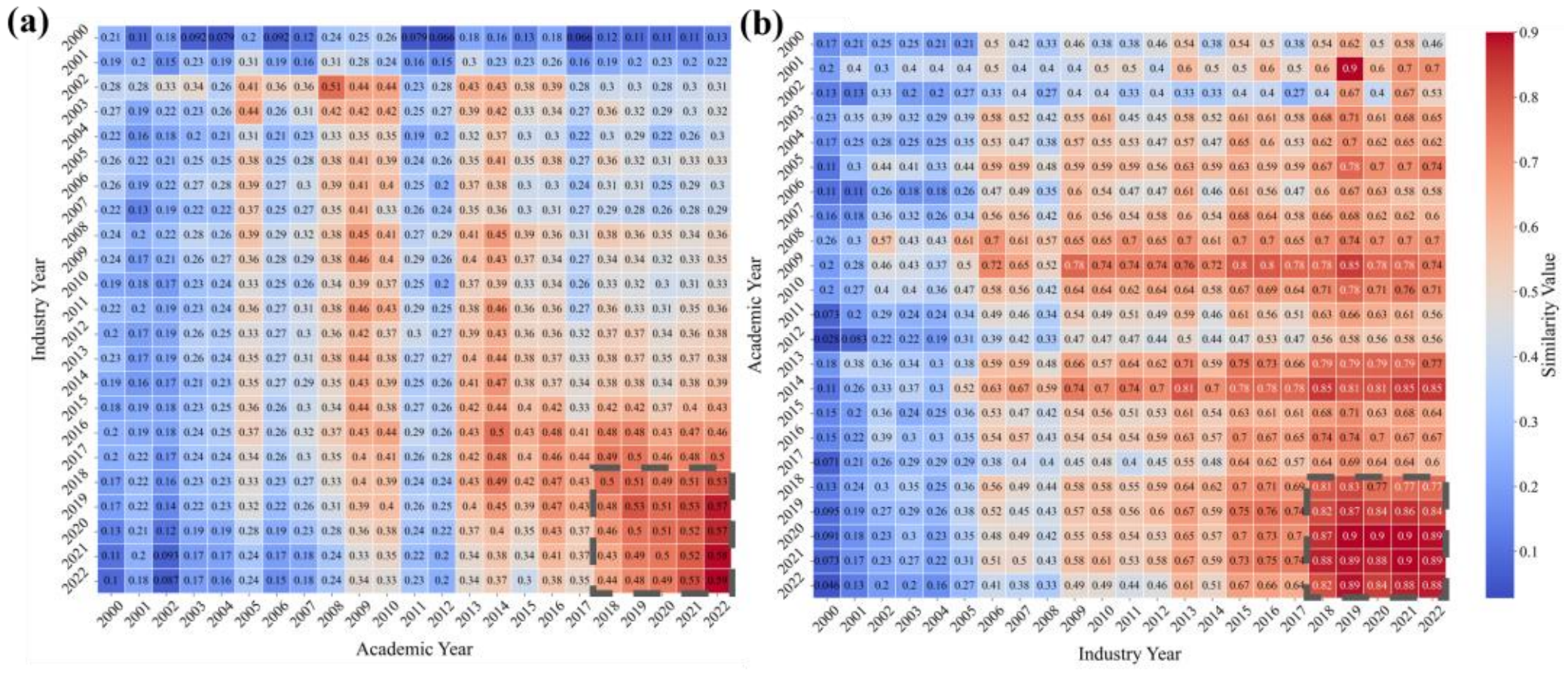

**Fig. 10.** Heatmap of semantic similarity. (a) The heatmap of similarity from academia searching for industry papers; (b) The heatmap of similarity from industry searching for academic papers.

*Notes.* The black dotted box in the lower right corner represents the pre-training paradigm stage after 2018, which is the brightest area in the heat map

Historically, IBM initiated machine translation research at the end of the last century, with its 1993 word-aligned translation model (Brown et al., 1993) marking the birth of statistical machine translation (SMT). Google officially launched the Beta version of Google Translate on April 28, 2006, which made statistical machine translation more accessible to more people. This trajectory exemplifies how technological innovations consistently offer new solutions to long-standing problems, particularly evident in machine translation. Around 2009, statistical methods and feature engineering remained dominant in NLP research. However, with the rise of deep learning (Bahdanau et al., 2016; Mikolov et al., 2013), especially after 2014, methods based on RNN proposed by academics gradually replaced traditional statistical models (Sundermeyer et al., 2014). This shift is not only an improvement in model performance but also greatly simplifies the complexity of feature engineering, enabling researchers to learn effective feature representations directly from raw data. These advancements propelled the field to higher levels and laid the foundation for the subsequent paradigm of self-attention mechanisms (Vaswani et al., 2017).

In addition, based on high similarity paper pairs, we matched the most relevant industry papers from previous years for each target academic paper, using 2009 and 2014 academic papers as the baseline reference, and generated word clouds as shown in **Fig. 11**. The high similarity topics are mainly centered around machine translation, and secondary topics include supervised learning and entity recognition. These word clouds characterize the closest early industrial research aligning with subsequent academic work, revealing new development opportunities that statistical machine learning and deep learning have generated for machine translation during technological advancement.

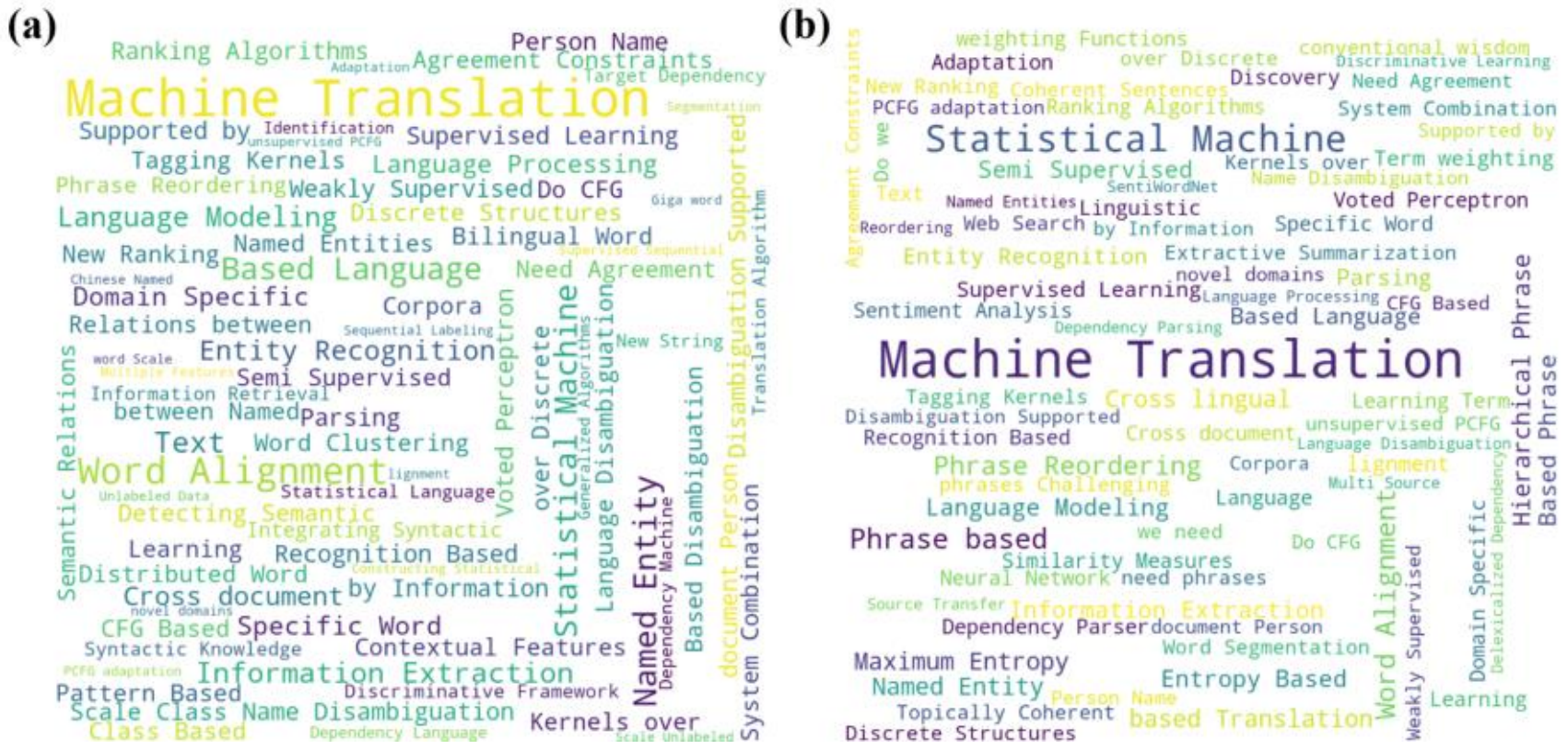

**Fig. 11.** Word cloud of industry papers identified as similar to 2009 and 2014 academic papers. (a) Word cloud of industry papers before 2009; (b) Word cloud of industry papers before 2014.

The research results not only reflect the pulling effect of industrial demand on the direction of academic research (Hoppmann, 2021), but also demonstrate the feedback that academic research brings to the practical problems of the industry, providing targeted solutions to industrial practice through theoretical innovation (Paniccia et al., 2025; Song & Wei, 2024). The resource advantages of the industry and the rich theories of academia form a synergy and jointly promote the development of the field.

## 4.2 Structural evolution of knowledge entity co-occurrence networks

In this section, we analyze the largest connected component, overlapping regions and modularity to understand the structure of knowledge networks in NLP.

### 4.2.1 The dominance of the knowledge network structure in academia gradually declines

We calculated the size and missing parts of the largest connected component from 2000 to 2022, as shown in **Fig. 12**. With the increase in publications and entities, the size of the largest component also grew annually. This reflects the rising complexity and scale of the network, indicating continuous knowledge development and integration in the NLP field.

In the early stages of NLP, due to limitations in technical theories, as well as restrictions in computational resources and power, the overall size of the entity network related to the technology was relatively small. During this period, academia dominated the knowledge network structure.

Since 2010, the rapid growth in dataset sizes, computational power (Strubell et al., 2019), and the rise of deep learning have driven breakthroughs in NLP. These changes are reflected in the increased role of industry and collaborations in the knowledge network structure, reducing academia’s dominance. After 2018, with the advent of pre-trained models, the knowledge network expanded rapidly, further highlighting the growing importance of industrial and collaborative contributions.

Academia's dominance has declined, partly due to the vast computational resources required for large models, which industry can better support. This also reflects NLP's increasing "knowledge burden" (Jones, 2009; Schweitzer & Brendel, 2021). With an increasingly complex body of knowledge, research has become more dependent on collaboration between academia and industry.

Furthermore, we draw the co-occurrence network of knowledge entities in different periods to visualize the trajectory of knowledge systems with industry and academia (as shown in **Fig. B3**).

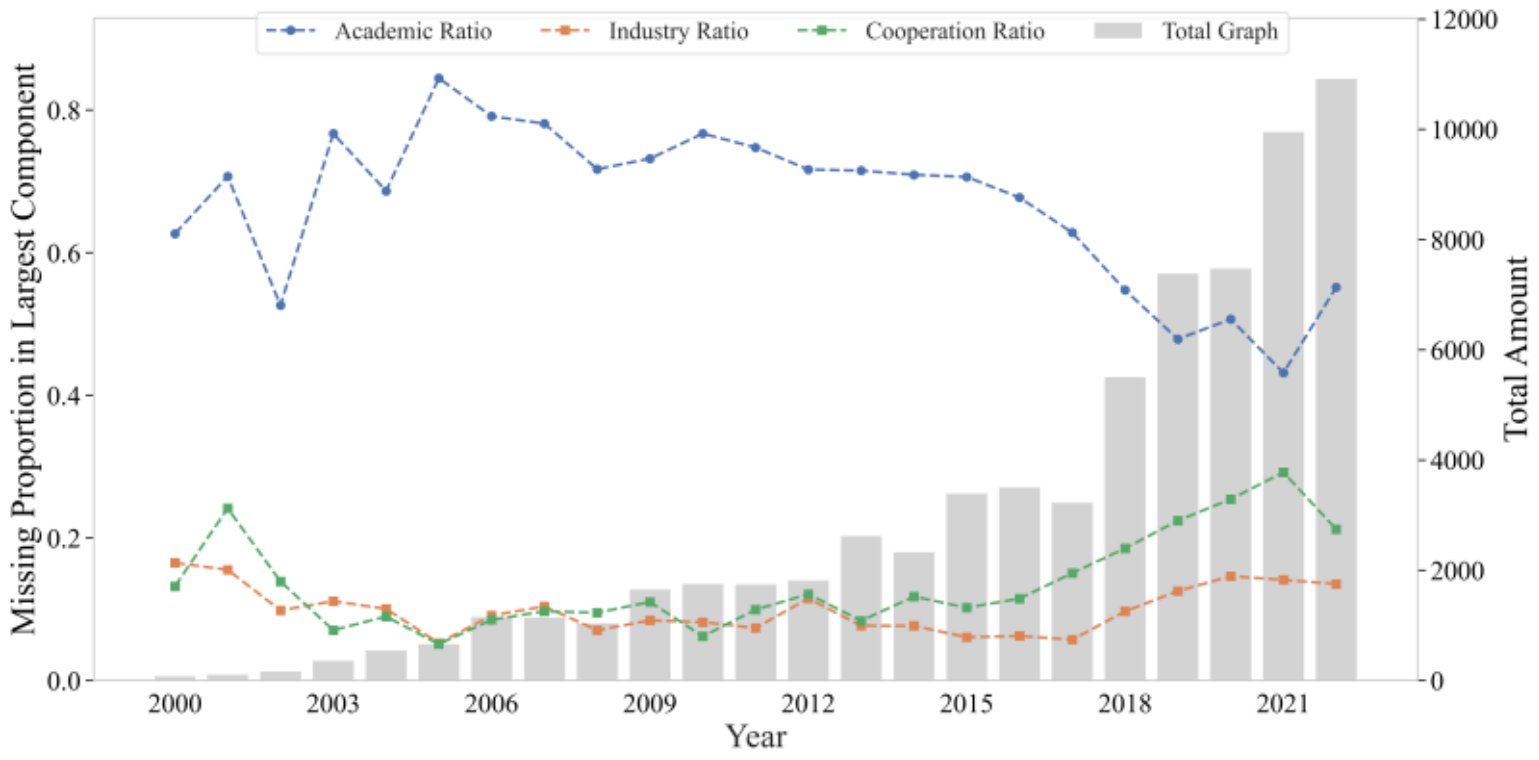

**Fig. 12.** Size of the largest connected component and the proportion of missing proportions by institution.

### 4.2.2 Topological characterization of knowledge entity network

We further explore the overlapping nodes of the largest connected component. Quantify the contribution of different types of institutions to knowledge reorganization by analyzing the characteristics of the connecting edges between public entities. The computational results are shown in **Fig. 13**.

We found that since the introduction of the pretrained model paradigm in 2018, the quantity of entity combinations has shown a significant increase. Additionally, the entity combination patterns among the three institutional types exhibited notable differences: the scale of entity linkage in academia is significantly higher than that in industry. This indicates that academia holds a greater advantage at the technical combination level, capable of constructing more extensive connections through common nodes in the knowledge network, thereby playing a pivotal role in knowledge dissemination and integration. Industry, on the other hand, consistently has the lowest portfolio size.

There are multiple mechanisms for the formation of this difference: First, the relatively low volume of publications in the industrial sector leads to its smaller representation in technical portfolios; Secondly, the technological innovation strategy of the industry tends to iterative optimization of existing technology rather than pursuing extensive knowledge exploration and combination. For instance, engineers often reduce innovation risk through module decoupling (Fleming & Sorenson, 2001). In contrast, academia, as an important source of knowledge in the innovation ecosystem (Rabelo Neto et al., 2024), has an advantage in terms of technology portfolio.

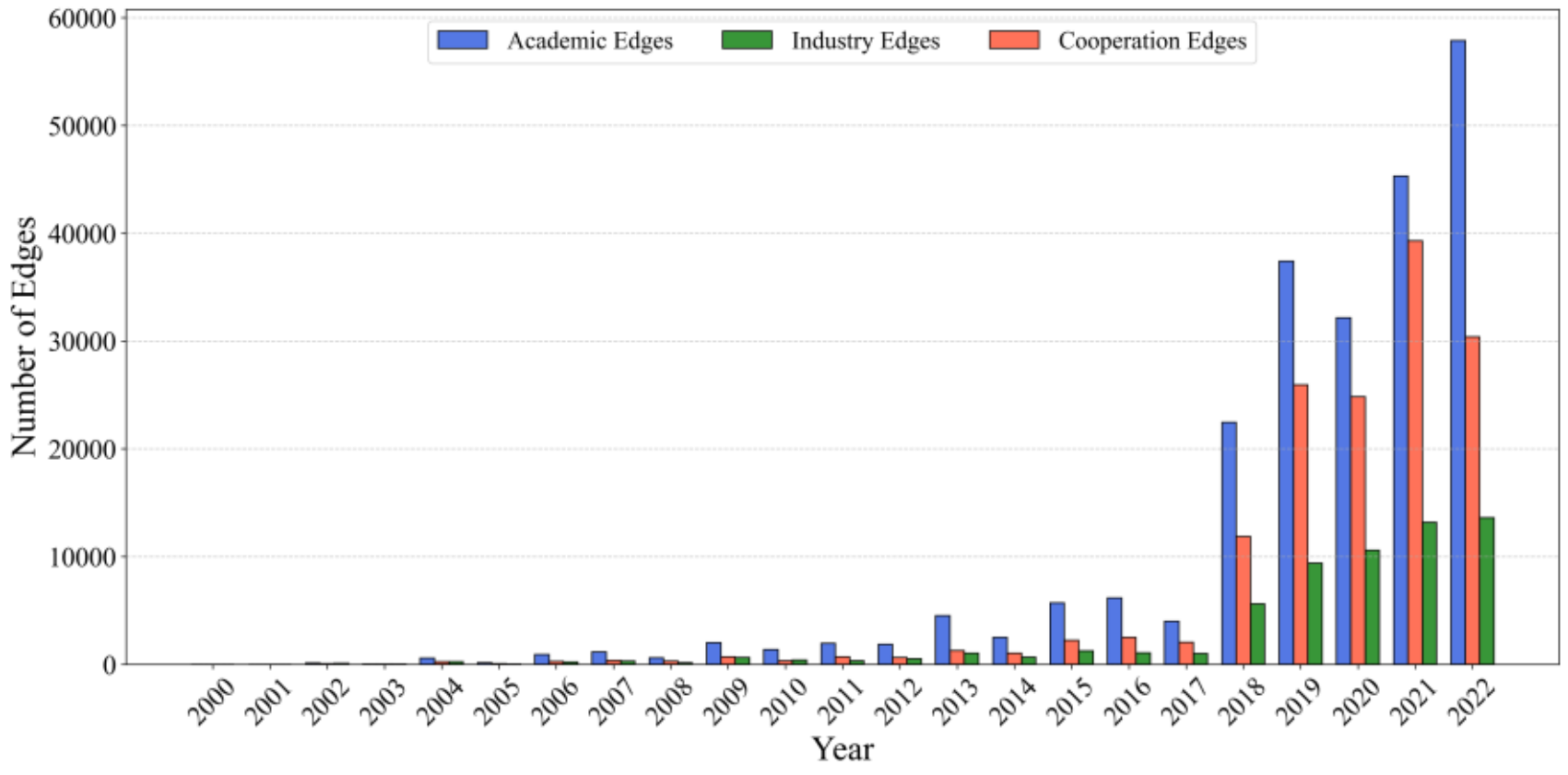

**Fig. 13.** Scale of edges formed between common nodes in subgraphs

**Fig. 14** presents the dynamics of the modularity of the network in the period 2000 - 2022 and categorizes it into three stages of technological development: statistical machine learning, deep learning, and pre-training. The change in modularity degree reveals the degree of integration of internal community groups in the co-occurring network of knowledge entities.

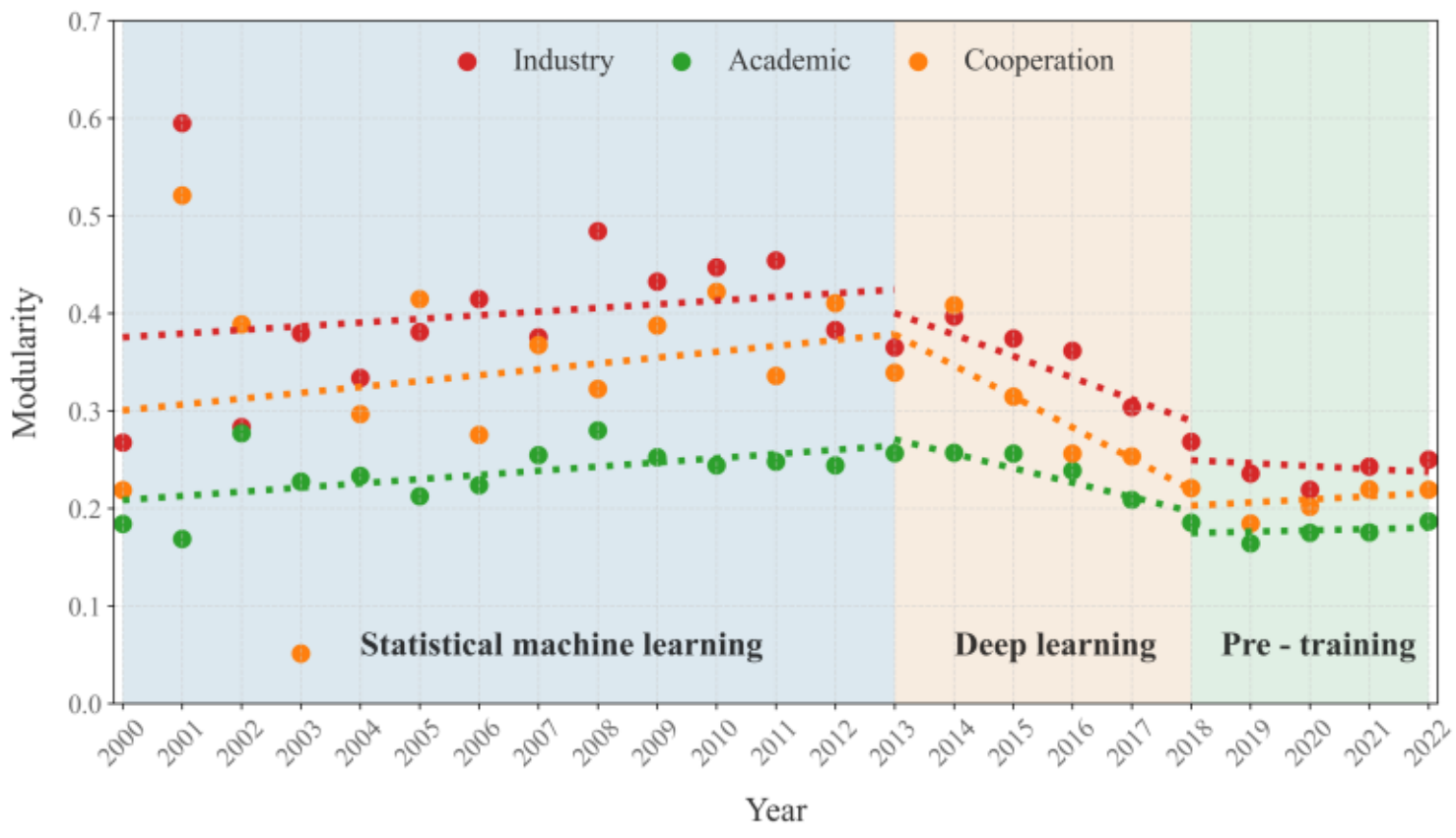

**Fig. 14.** Trend of modularity change in different subgraphs

In the statistical machine learning stage, the modularity of the industry continues to be higher than that of the academic and cooperative institutions, and the value stays above 0.3 and shows an upward trend, which further indicates that a strong community structure and knowledge integration tightness is maintained within it, with limited communication between communities. This phenomenon reflects the industry's high degree of focus on specific applications (e.g., machine translation), leading to a path dependency and lock-in effect in its body of knowledge (Colombelli et al., 2021). In contrast, the modularity of academia is low and remains stable, indicating that its network structure is more open and that knowledge dissemination and integration remain in a steady state.

In the deep learning era, the modularity of industry and collaborative institutions decreases significantly, signaling that technological progress have broken through tightly knit communities within the field and facilitating integration between different communities in the knowledge network. During the pre-training stage, the modularity of the three types of institutions stabilized and remained at a low level.

Overall, the change in the degree of modularity reveals the characteristics of knowledge integration and diffusion, and this change is essentially an outward manifestation of the synergistic evolution between technological development and institutions. Technological developments drive institutions from relatively independent communities to open and interactive network structures.

## 4.3 Changes in the direction of knowledge flow between industry and academia

In this section, we address the question RQ3 by measuring the knowledge flow in the NLP field through references.

### 4.3.1 Knowledge flow becomes more bidirectional between industry and academia

We calculate the proportion of different types of citations by comparing the frequency of references from specific institutions to the total frequency of all references. For example, to assess the knowledge inflow from industry to academia in 2018, we measured this using the ratio of the number of times industry cited academic articles in 2018 divided by the total number of industry citations in that year.

**Fig. 15** illustrates that academic citations represent the highest proportion among all citations. Initially, academia was the main contributor to foundational algorithms and theories in NLP. All three publication types primarily absorbed insights from the industrial sector, with over half of the citations consistently sourced from academia across all periods.

From 2000 to 2012, the proportion of citations from academia increased, while citations to industry declined. This period indicated that academia was the primary knowledge source in statistical machine learning. However, since 2013, the rise of deep learning as a data- and computation-driven AI subfield has shifted the balance toward industry, increasing its citation proportion and a decrease in academic citations. Although academia remains a major knowledge source, industry has significantly changed the knowledge flow dynamics, leveraging advantages like computational power, large datasets, and skilled researchers (Ahmed et al., 2023). Concurrently, the number of collaborative publications has risen (see **Fig. 6**), with citation frequencies increasing rapidly (Färber & Tampakis, 2024), fostering more active cross-disciplinary collaboration and enhancing knowledge flow.

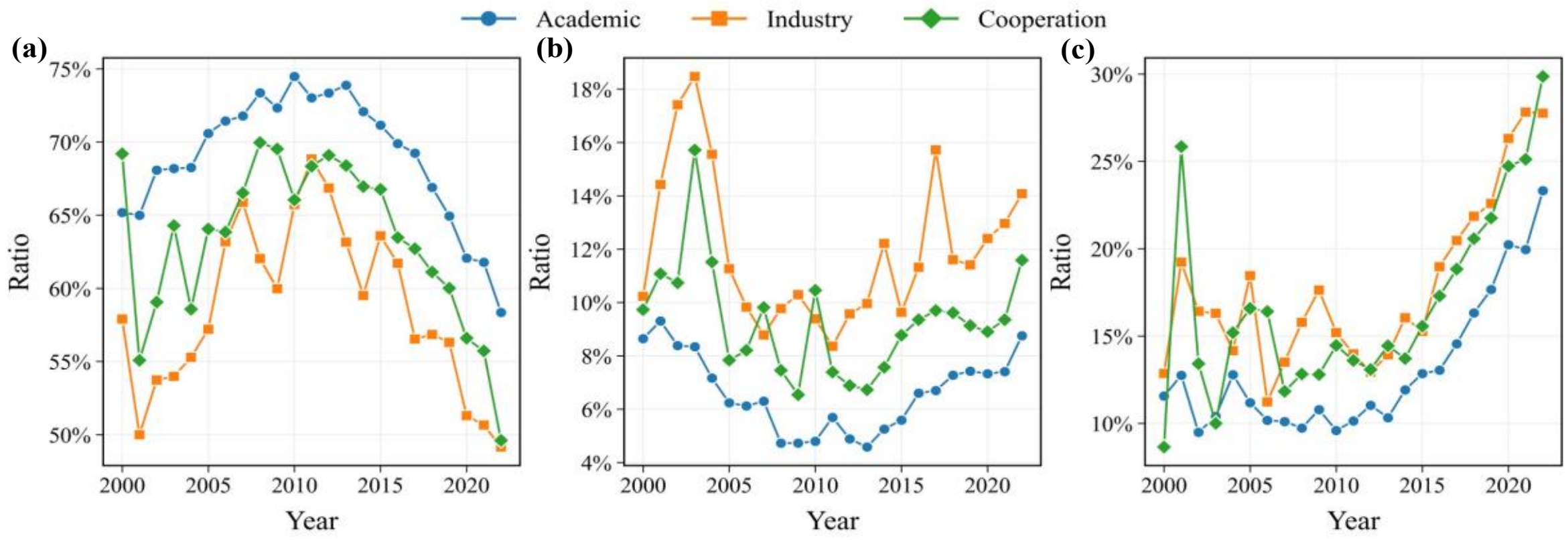

**Fig. 15.** Different types of citations and their distribution: (a) Proportion of citations from academic papers by different types of publishing institutions; (b) Proportion of citations from industry papers by different types of publishing institutions; (c) Proportion of citations from cooperation papers by different types of publishing institutions.

In 2017, the emergence of the Transformer model triggered a peak in internal citations within the industrial sector, while the proportion of citations from industry to academia decreased, a trend that continues to this day. Although the citation proportion from academia has declined, this does not imply a reduction in its influence. Academic citations still dominate, with all three types of publications citing academia at rates exceeding 50%. Overall, the proportion of knowledge flowing out of industry has gradually increased, strengthening the bidirectional flow of knowledge in the NLP field. Notably, the influence of collaborative publications continues to rise, accounting for approximately 30% of total citations, becoming a significant source of knowledge.

### 4.3.2 Bidirectional knowledge flow correlates with increased knowledge proximity

To quantify the direction of knowledge flow and the degree of proximity between academia and industry, thereby assessing their potential correlation, we first conducted a macro-level analysis by calculating the HHI of annual citation distributions and compared it with the entity similarity across different years.

As shown in **Fig. 16**, this suggests that a more even distribution of citations facilitates greater communication among research entities. The bidirectional flow of knowledge within the field effectively promotes research collaboration and the dissemination of outcomes, further enhancing the alignment of research content between academia and industry.

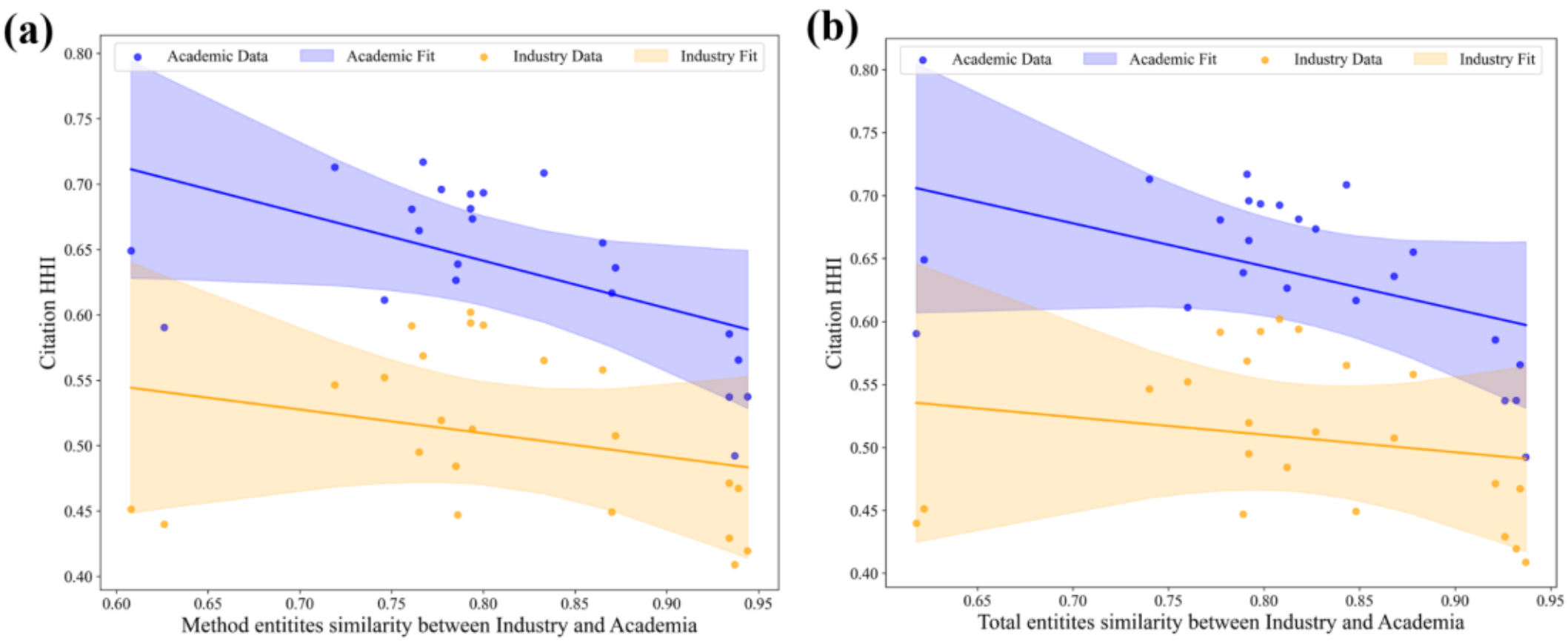

**Fig. 16.** Association between entity similarity and citation distribution HHI. (a) Only use the method entity; (b) Use all entities.

To further analyze the research question, we use a multiple regression model constructed in Section 3.4.2 to examine the correlation between the concentration of citation distribution HHI and similarity for each paper. To further reveal the impact of technological evolution on this pattern, we still build regression models according to each of the three key phases in the NLP field (2000–2012 statistical machine learning phase, 2013–2017 deep learning phase, and 2018–2022 pre-training model phase).

Descriptive statistics and correlation coefficients of the variables are shown in **Table 5** and **Fig. 17**, where the correlations between the independent variables do not indicate potential multicollinearity. For more robust results, we performed multicollinearity tests before multiple regression. The variance inflation factor (VIF) of all independent variables is less than 2, and the average VIF is 1.34, indicating no serious multicollinearity. The final regression results are shown in **Table 6**, and the key findings are as follows.

Overall, in the base model Model (1) with only HHI and fixed effects, HHI has a significant negative effect on semantic similarity ($\beta = -0.037, p < 0.01$), suggesting that the more evenly distributed the citations in a paper are (the lower the HHI), the more closely it is associated with industry or academia. Model (2) remains significant ($\beta = -0.032, p < 0.01$) after adding the control variables (logarithm of number of institutions, authors, and references).

Subsequently, we analyzed each of the three periods. In the statistical machine learning phase (2000–2012), HHI had a significant negative effect on semantic similarity ($\beta = -0.021, p < 0.01$). However, after adding control variables in Model (4), the coefficient of HHI continues to become smaller ($\beta = -0.014, p < 0.01$), suggesting that the effect of the distribution of citation concentration may be partially explained by other factors. In addition, the relatively small coefficients suggest that knowledge flow during the statistical machine learning stage contributes limitedly to the improvement of knowledge proximity.

In the deep learning phase, the balanced distribution of citations enhances semantic similarity ($\beta = -0.051, -0.045, p < 0.01$) regardless of the inclusion of control variables, as shown in Model (5) and Model (6). During the pre-training model phase, the negative effect of HHI persisted in Model (7) and Model (8) ($\beta = -0.040, -0.043, p < 0.01$). All models include fixed effects for year and institution type, excluding time trends and institution type. R-squared improved with controls in all cases, suggesting that adding control variables to the model improves the goodness of fit.

**Table 5** Descriptive statistics of variables (N = 13,225).

| *Variable* | *Mean* | *Std. Dev.* | *Min* | *Max* |
|---|---|---|---|---|
| SemSim | 0.724 | 0.036 | 0.464 | 0.809 |
| HHI | 0.647 | 0.176 | 0.333 | 1 |
| Institution_num | 1.542 | 0.891 | 1 | 20 |
| Author_num | 3.408 | 1.861 | 1 | 37 |
| References | 25.094 | 14.920 | 1 | 97 |
| Institution type | 0.895 | 0.307 | 0 | 1 |

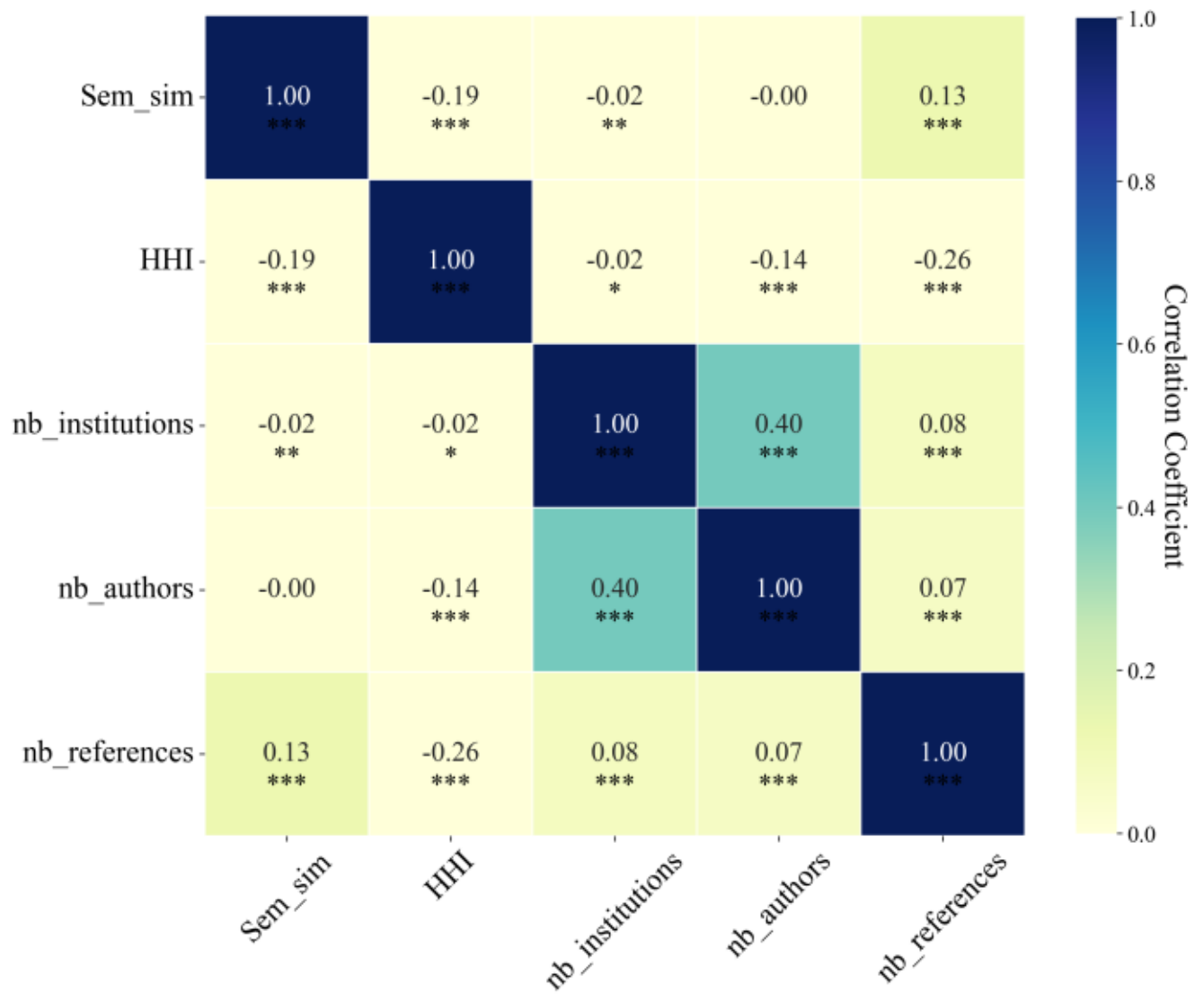

**Fig. 17.** Correlation coefficient between variables

The correlation reflected in the final regression reflects the integration effect formed in the bidirectional flow of knowledge: a uniform citation distribution implies a more balanced dissemination of knowledge among different subjects, which promotes the fusion of knowledge systems.

**Table 6** The association between citation distribution HHI and semantic similarity.

| | Overall Period (2000–2022) | | Statistical machine learning (2000–2012) | | Deep learning (2013–2017) | | Pre-training (2018–2022) | |
|---|---|---|---|---|---|---|---|---|
| VARIABLES | **Model(1)** | **Model(2)** | **Model(3)** | **Model(4)** | **Model(5)** | **Model(6)** | **Model(7)** | **Model(8)** |
| *HHI* | -0.037*** | -0.032*** | -0.021*** | -0.014*** | -0.051*** | -0.045*** | -0.040*** | -0.043*** |
| | (0.002) | (0.002) | (0.004) | (0.004) | (0.005) | (0.005) | (0.003) | (0.003) |
| *ln(nb_institutions)* | | -0.003*** | | -0.002 | | -0.003 | | -0.002** |
| | | (0.001) | | (0.002) | | (0.002) | | (0.001) |
| *ln(nb_authors)* | | -0.000 | | 0.001 | | -0.002 | | -0.002* |
| | | (0.001) | | (0.001) | | (0.002) | | (0.001) |
| *ln(nb_references)* | | 0.003*** | | 0.013*** | | 0.005*** | | -0.001** |
| | | (0.000) | | (0.001) | | (0.001) | | (0.001) |
| *_cons* | 0.722*** | 0.713*** | 0.710*** | 0.673*** | 0.750*** | 0.732*** | 0.752*** | 0.761*** |
| | (0.005) | (0.005) | (0.005) | (0.006) | (0.004) | (0.006) | (0.002) | (0.004) |
| Year fixed effects | Yes | Yes | Yes | Yes | Yes | Yes | Yes | Yes |
| Institution type effects | Yes | Yes | Yes | Yes | Yes | Yes | Yes | Yes |
| Observations | 13,225 | 13,225 | 3,726 | 3,726 | 3,002 | 3,002 | 6,497 | 6,497 |
| R-squared | 0.036 | 0.050 | 0.022 | 0.065 | 0.045 | 0.054 | 0.036 | 0.039 |

Robust standard errors in parentheses，*** p<0.01, ** p<0.05, * p<0.1

### 4.3.3 Analysis of citation behavior differentiation

Meanwhile, as observed in **Fig. 15**, the distinct citation preferences between academia and industry in their respective research. Academia demonstrates a stronger inclination to cite papers within the academic domain, whereas industry tends to favor citations from industrial publications. This citation pattern reflects the divergent research priorities and objectives of the two sectors: academia typically emphasizes theoretical exploration, methodological innovation, and scholarly discourse, while industry is more focused on practicality, application outcomes, and addressing market-driven needs.

To further explore this phenomenon, we analyzed the excess self-citation between academia and industry to measure the extent of their attention to each other's research. **Fig. 18** shows the trend between academia and industry at the level of self-citation rate, where values greater than zero represent a preference for self-citation and values less than zero represent cross-citation.

It can be observed that industry shows a stronger tendency to cite articles from the same institutional type compared to academia. Academic researchers are relatively less likely to self-cite when citing, meaning that they are more likely to be familiar with both their own work and that of industry, and that academic research covers a wider range of content (Liang et al., 2024).

Examining **Fig. 18** further, we find that there has been a certain increase in the excess self-citation rate in academia since 2016. This phenomenon may be closely related to the growth of the model's computational demand, suggesting a structural adjustment in the research orientation of academia. In particular, when research topics are under pressure from high computational power consumption and hardware costs, the pattern of resource allocation and the degree of focus of academics in specific research directions may have changed subsequently.

In exploring the situation of ECC rates in academia after 2016, we consider that there is a high dependence on GPUs for fine-tuning deep learning and pre-trained models. In addition, hardware-related terms such as GPU, TPU, CUDA, CPU, etc., are closely related to cutting-edge distributed computing tasks such as deep learning and pre-trained models. We use these hardware-related terms as indicators, assuming that papers mentioning them directly exhibit higher computing resource dependency for distinguishing studies with varying computational demands.

We obtained 642 papers (including 512 academic papers) containing the above entities, which we categorized as high computational demand papers. The rest are categorized as low computational demand research papers. The Mann-Whitney U test found that the academic citations of low computational demand papers are significantly higher than those of high computational demand papers ($p < 0.001$). This result suggests that high computational demand research may rely more on the accumulation of technical practices and pay relatively less attention to the innovations of academia.

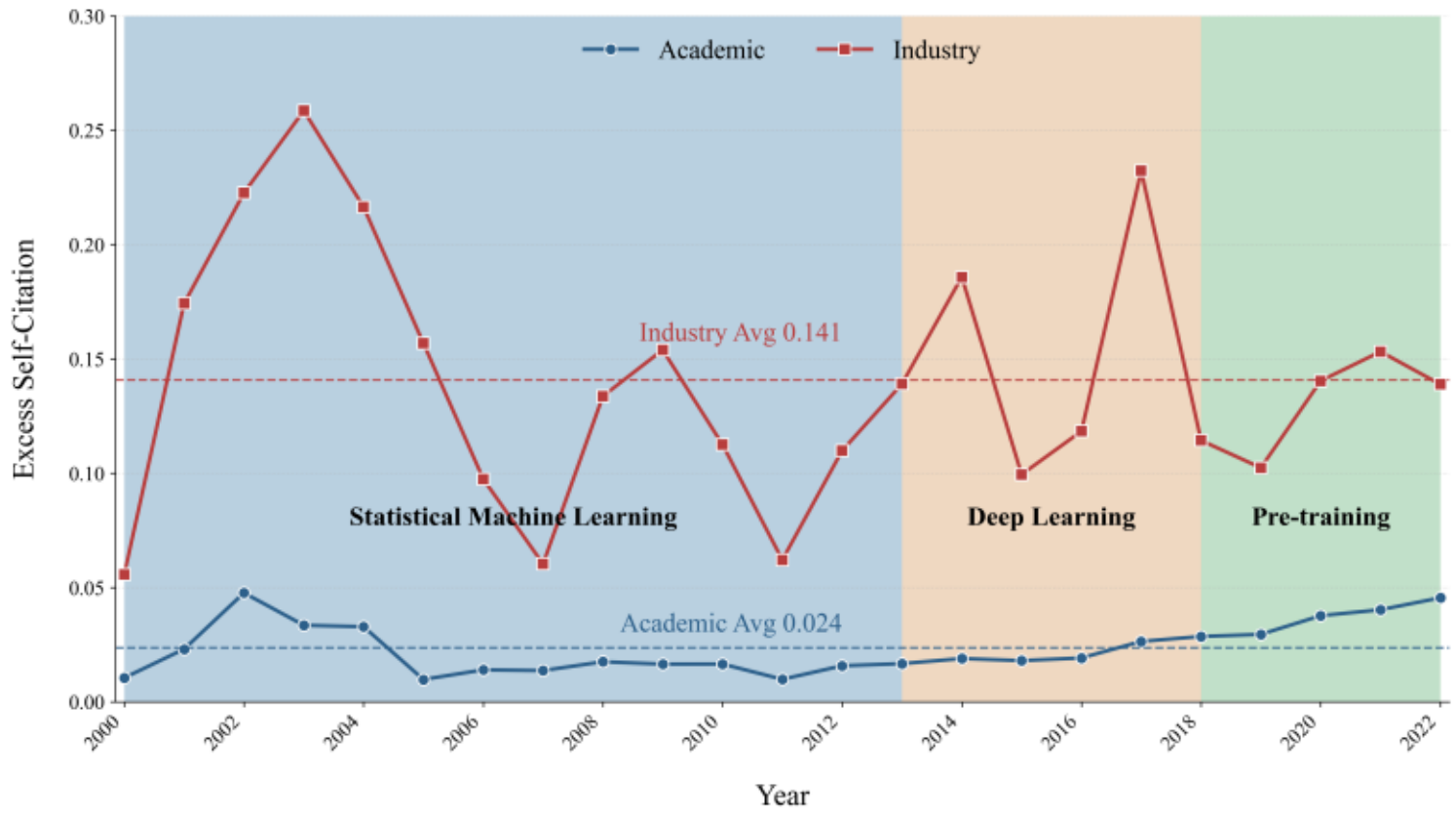

**Fig. 18.** Excess self-citation in academia and industry

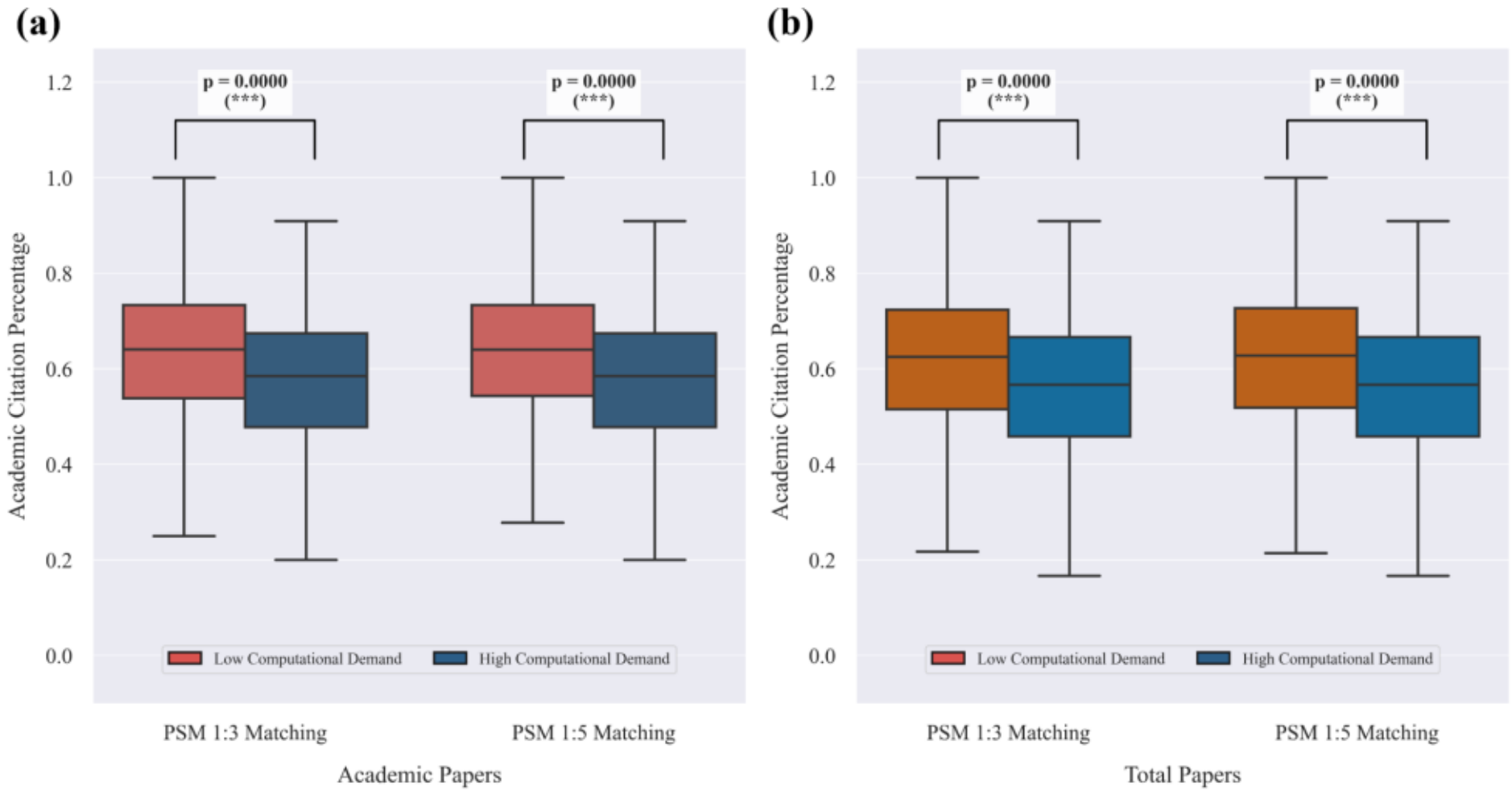

**Fig. 19.** Boxplot of the proportion of academic papers cited by different computing demands.

*Notes.* ***: p<0.001, **: p<0.01, *: p<0.05.

To prevent potential bias, we further employ propensity score matching (PSM) for robustness testing. We eliminated covariate differences between different computational demand groups by estimating the propensity scores through a Logit model using the number of institutions, number of authors, number of references, and year as covariates. A balanced control group sample was constructed using 1:3 and 1:5 nearest neighbor matching strategies for PSM, respectively, and the results are shown in **Fig. 19**. Both for academic papers and all types of papers, the citation ratio of high computational demand papers to academia is lower than that of low computational demand papers. Conversely, low computational demand studies cite academic papers more frequently.

# 5 Discussion

**Table D1** summarizes key findings of this study. Drawing on these, we elaborate on their complementarity and extensions to existing literature on academia-industry links.

Firstly, this study quantitatively analyzes from the perspective of fine-grained knowledge. This study reveals a steady rise in industry-academia knowledge proximity over time, with the 2018 emergence of pre-training paradigms significantly accelerating knowledge convergence. Core technologies like Transformer and BERT rapidly proliferated due to their superior performance, becoming shared knowledge carriers across sectors. Meanwhile, open-source ecosystem development broke inter-institutional knowledge barriers, enhancing the openness and accessibility of NLP technologies and methods. Further, heatmap analysis intuitively illustrates discontinuous technological change (Dosi, 1982; Kuhn, 1962) (see **Fig. 8**). At the technical standard level, **Fig. B1** (c) indicates strong alignment in Metric entities between sectors, standardization of core elements like evaluation metrics laid the groundwork for knowledge sharing (Aufrant, 2022), enabling researchers to advance toward unified goals.

Second, alongside technological development and rising knowledge proximity, two key shifts emerged in knowledge network structures. (1) academia's dominance weakened, with more balanced distribution breaking early single-dominance; (2) network modularity declined, as closed internal knowledge community boundaries between sectors dissolved, integrating more heterogeneous knowledge into the NLP network. This restructuring stems from multiple factors. The growing knowledge burden in NLP (Jones, 2009; Schweitzer & Brendel, 2021), requires cross-sector collaboration. Resource dependency has shifted, with industry dominating the supply of core factors (Ahmed et al., 2023). At the same time, the industry has also actively integrated into academic systems, participating in top conference submissions (Färber & Tampakis, 2024), driving knowledge openness.

Third, this study validates and extends knowledge flow theory, confirming mutual reinforcement between knowledge proximity and knowledge flow. Higher proximity reduces inter-institutional knowledge flow barriers, while more frequent flows further narrow the academia-industry knowledge proximity (Chen, Mao, Ma, et al., 2024; Dolfsma & van der Eijk, 2016; Sudhindra et al., 2020; Syafiandini et al., 2024). The study also finds that resource dependency significantly influences knowledge flow. Research with high computational power requirements will be more closely associated with industry due to its reliance on hardware resources from the industry, while reducing its dependence on academia.

## 5.1 The impact of external context on academia and industry knowledge proximity

Although this study focuses on mining textual evidence of changes in knowledge proximity between academia and industry, external environments such as policy support, market trends, and corporate strategies also have multilevel impacts on knowledge proximity between academia and industry. We discuss these factors in this section.

At the institutional and legal level, the Bayh-Dole Act of 1980 in the United States, by clarifying the intellectual property ownership of publicly funded research results (Chai & Shih, 2016; Mowery et al., 2001), reduced the transaction costs of knowledge transfer and promoted the emergence of the third mission of universities (Chai & Shih, 2016). It has provided an effective institutional incentive for tripartite cooperation among government, academia, and industry to work together on the commercial utilization of government-funded R&D results. European academia-industry partnership programs, such as the Fraunhofer-Gesellschaft and the Danish National Advanced Technology Foundation, practice empowering public research institutions to support enterprise innovation. At the same time, these policies have spawned the widespread establishment of joint laboratories, as well as intermediary institutions such as TTOs, which provide organizational vehicles for academia-industry knowledge matching (Meissner et al., 2022).

From the perspective of market trends and corporate strategy, the Apache Software Foundation[9] and Huggingface community[10] represent decentralized knowledge production networks. By opening up technical frameworks, they reduce industry barriers and attract academic institutions to engage in model development. However, the dominance of the head firms in core resources such as computing power, large datasets, and talent (Ahmed et al., 2023) makes it possible to maintain their technological advantage by controlling the supercomputing clusters required for distributed training even if they open-source the underlying model.

Funding is also a crucial factor affecting proximity. Obtaining corporate funding can provide students with new learning opportunities, thereby promoting their research activities (Lee, 2000). At the same time, this will also lead to a reorientation of research in academia towards a gradual shift towards practice-oriented research, which in turn leads to an application-driven model of knowledge production (Hoppmann, 2021) and promotes close synergies. Yet there are drawbacks to a high reliance on industry funding. This can lead to a risk of technological lock-in as the direction of the technology is deeply tied to the needs of the industry (Hoppmann, 2021). Once the industrial demand changes or the technological development encounters bottlenecks, the vulnerability in the technological system will be exposed.

---

[9] https://apache.org/

[10] https://huggingface.co/

## 5.2 Implications

We explain the significance of this paper from both theoretical and practical perspectives.

### 5.2.1 Theoretical implications

Research has revealed the dialectic between institutional logic and knowledge production (Vallas & Kleinman, 2007). The new paradigm brought about by innovative technologies in NLP (Kuhn, 1962), a scientific revolution (e.g., the Transformer architecture), has led to a reconfiguration of the knowledge system. This has increased the knowledge proximity between industry and academia

The feedback effect of technological change on the knowledge ecosystem is vividly illustrated in this study: Take the machine translation task (Brown et al., 1993) as an example, which was initially a key application scenario prioritized by the industry. It not only validates the pulling effect of industrial needs on the direction of academic research (Hoppmann, 2021), but also demonstrates the feedback that academic research brings to practical problems in industry (Paniccia et al., 2025; Song & Wei, 2024). This has enabled machine translation to regain vitality after many rounds of technological changes, such as statistical learning and neural networks, and has become a landmark example of academia-industry collaboration in NLP.

In terms of the dynamic evolution of knowledge dominance, this study presents the trajectory of knowledge dominance reconstruction. Whether observed from the knowledge co-occurrence network or the direction of knowledge flow, academia dominated the knowledge in the early stage. However, along with the continuous development of technology, the dominance of academia gradually decreases, and the bidirectional flow of knowledge is characterized by an increase. At the same time, the exponential growth of domain knowledge also raises the issue of “knowledge burden” (Jones, 2009; Schweitzer & Brendel, 2021), highlighting the need for academia-industry collaboration to cope with complexity.

At the methodological level, this study adapts text mining and bibliometrics methods to innovation theory, breaking through the limitations of traditional macro indicators (Aufrant, 2022; Blankenberg & Buenstorf, 2016; D’Este et al., 2019; Fischer et al., 2019; Powell et al., 2005). This innovation not only provides a quantitative tool for NLP, but also provides a replicable analysis paradigm for other subfields of AI, and technology-intensive fields such as biomedicine and semiconductor technology by adapting the theory and methodology.

### 5.2.2 Practical implications

Our research shows that technological paradigm change and knowledge flow are closely synergistic. The pre-training paradigm significantly improves knowledge proximity. Subsequently, an open technology sharing platform should be actively constructed, drawing on open-source collaboration, to establish a cross-institution sharing platform covering core elements such as technical methods, tools, and datasets, and to shorten the transformation cycle of laboratory results to industrial applications. Achieve the integration of cutting-edge methods in academic research and industrial engineering tools, facilitate bidirectional knowledge element flow via standardized interfaces and cut repetitive R&D costs.

For university TTOs, can develop an intelligent docking system based on technology entity matching, adopt open-source protocols for sharing basic theory results, and patent cross-licensing for engineering tools, balancing academic openness and industrial commercialization needs. Crack the problems of high transaction costs and low matching efficiency in traditional technology transfer (Zaini et al., 2018). In addition, other fields can also learn from the model of open source community and open source ecology in NLP, to promote enterprises to share tools and datasets in technical practice in the form of open licenses, while academics transform theoretical achievements into reusable technical modules through joint research.

Given that industry may dominate in AI research (Färber & Tampakis, 2024), the government needs to play a key role at the regulatory level in developing a comprehensive and operational AI research code. At the same time, governments can also promote the involvement of academia in the monitoring system of AI research. The government can provide more external support to academia, covering key elements of AI research such as hardware facilities, computing power, and data resources. This will help academia continue to shape the frontiers of modern AI research, balancing the relationship between industrial interests and public well-being (Ahmed et al., 2023). Moreover, academic evaluation can be improved within the academic community. Increase the recognition of research results that engage in basic AI research with a focus on the ethical impact on society, and by optimizing the evaluation criteria, guide academic research toward a diverse, public interest-oriented development, and motivate scholars to conduct more socially responsible research.

## 5.3 Limitations

This study has several limitations. First, by focusing on English language top conferences, the data sources were restricted to papers and citation data from the three major conferences: ACL, EMNLP, and NAACL. This ignores other conferences and journals in the NLP field, as well as patent data. This may underestimate the innovative contributions of industrial institutions in technological application (Sauermann & Stephan, 2010). However, considering the characteristics of the NLP discipline, where cutting-edge knowledge is primarily disseminated through top conferences, this study's characterization of knowledge within the NLP field remains representative and can effectively capture the evolution trajectory of mainstream paradigms in knowledge production.

Second, due to the "multidirectional causality" in the NLP ecosystem, we recognize that industry and academic driver explanations are limited in their ability to explain causality at the causal level, and that it is difficult to strip the causal effect of a single variable through exogenous shocks. As Lewin and Volberda (1999) mention, there are multidirectional causalities to this co-evolution. The study by Abatecola et al. (2020) also suggests that relationships in co-evolution are reciprocal and simultaneous. Currently, in this study, which side of the industry and academia dominates the other? This question remains unanswered.

At the data level, the OpenAlex data used in this study, while providing a rich resource for academic research, may also introduce certain biases. For example, OpenAlex data may have missing information about institutions (Zhang, Cao, et al., 2024), or incomplete reference inclusion. These may have an indirect impact on the study results.

Finally, there may be an impact of bandwagon effects in the research (Zuckerman, 1977). When Transformer and the pre-training paradigm became research hotspots, a large number of studies may have focused on them to follow the trend. Particularly for certain studies with limited innovativeness, their potential influence on research conclusions still requires more exploration.

# 6 Conclusion and future works

This study quantified the similarity of research content between academia and industry from both the entity level and the semantic level, integrating citation evidence. It systematically reveals the coevolutionary trajectories of knowledge production paradigms between academia and industry from 2000 to 2022.

In future work, first regarding data sources, the current study only uses conference papers from three major conferences for research in the NLP field. We have not utilized data such as patent documents, journal articles, and foundation project texts. In subsequent research, incorporating these additional data types will enable more comprehensive analysis.

Second, as the link of the innovation system (Etzkowitz & Leydesdorff, 2000), there is still room for expansion in the quantitative analysis of external policy and institutional environments in this study. External factors such as government research funding, industrial policy orientation, and intellectual property rights systems, as well as resource allocation logics, influence academic-industrial research directions. However, the current study has not yet fully incorporated the impact of these policy variables on knowledge production. For example, the extent to which government funding for NLP technologies influences the consistency of academic-industrial technological pathways still requires further investigation in subsequent research.

Finally, at the level of causal interpretation, causal inference tools can be further utilized in the future or based on technological event-driven case studies. Taking key time nodes such as the rise of the pre-training paradigm (2018) and the ChatGPT technological breakthrough (2022) as entry points, specifically analyze the academia-industry characteristics before and after these years.

# Appendix A. The descriptions and examples of entities.

**Table A1** The descriptions and examples of entities.

| Type | Descriptions (Luan et al., 2018; Zhang, Zhang, et al., 2024) | Examples |
|---|---|---|
| Method | Algorithms, models, frameworks, etc | BiLSTM, RNN, Attention, BERT, Transformer |
| Dataset | Data, resources, Corpus, Knowledge base | WordNet, Wikipedia, TreeBank, Twitter, SQuAD |
| Tool | Programming languages, software, open-source tools, etc | Python, Pytorch, Spacy, GIZA++, TensorFlow |
| Metric | Measures that can express quality of a method | Accuracy, BLEU, Recall, ROUGE, Cross-Entropy |

# Appendix B. Visualization experiment

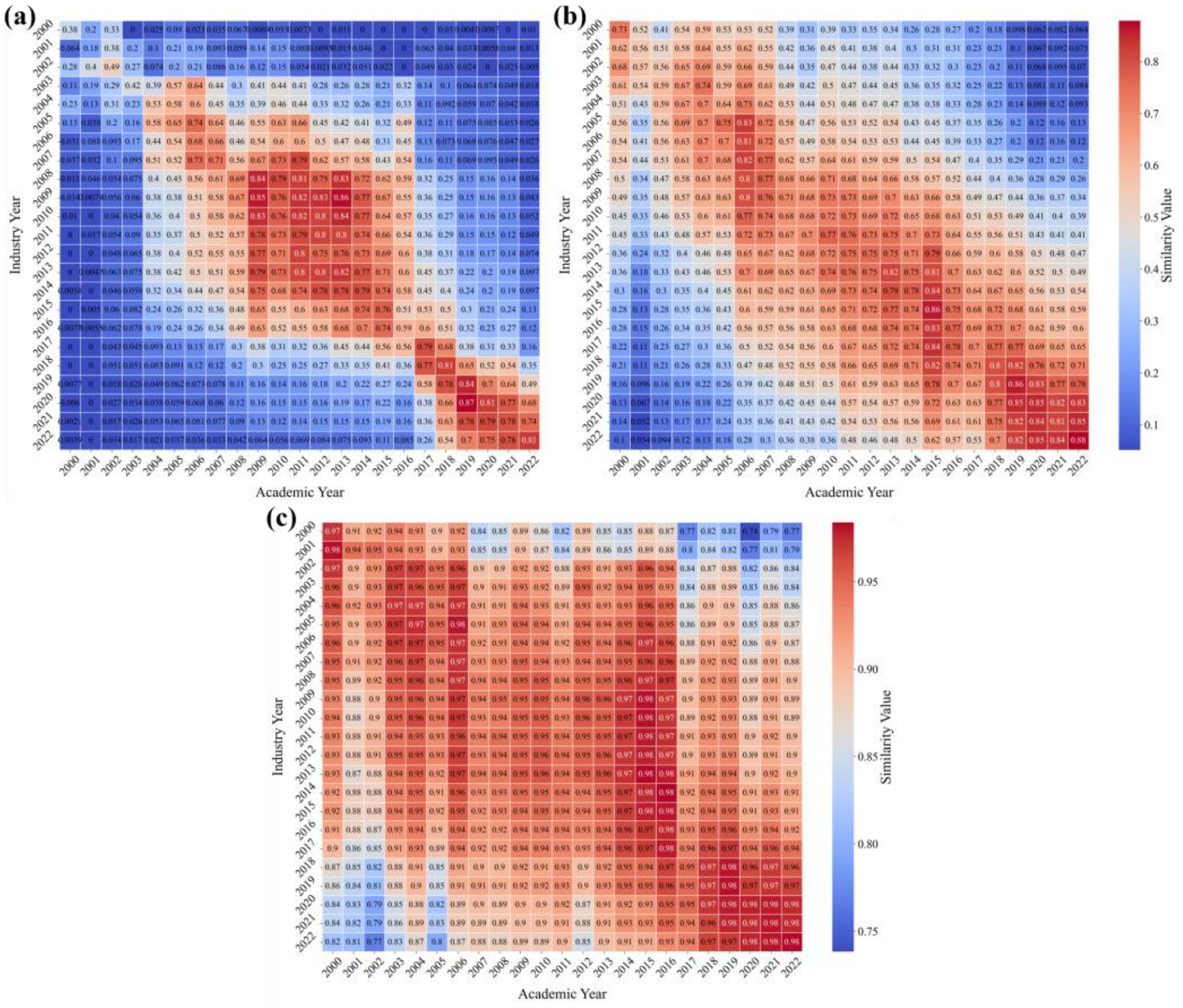

**Fig. B1.** Heatmap of bag-of-words vector similarity for entities. (a) The heatmap of bag-of-words vector similarity is based on Tool entities; (b) The heatmap of bag-of-words vector similarity is based on the dataset entities; (c) The heatmap of bag-of-words vector similarity is based on the metric entities;

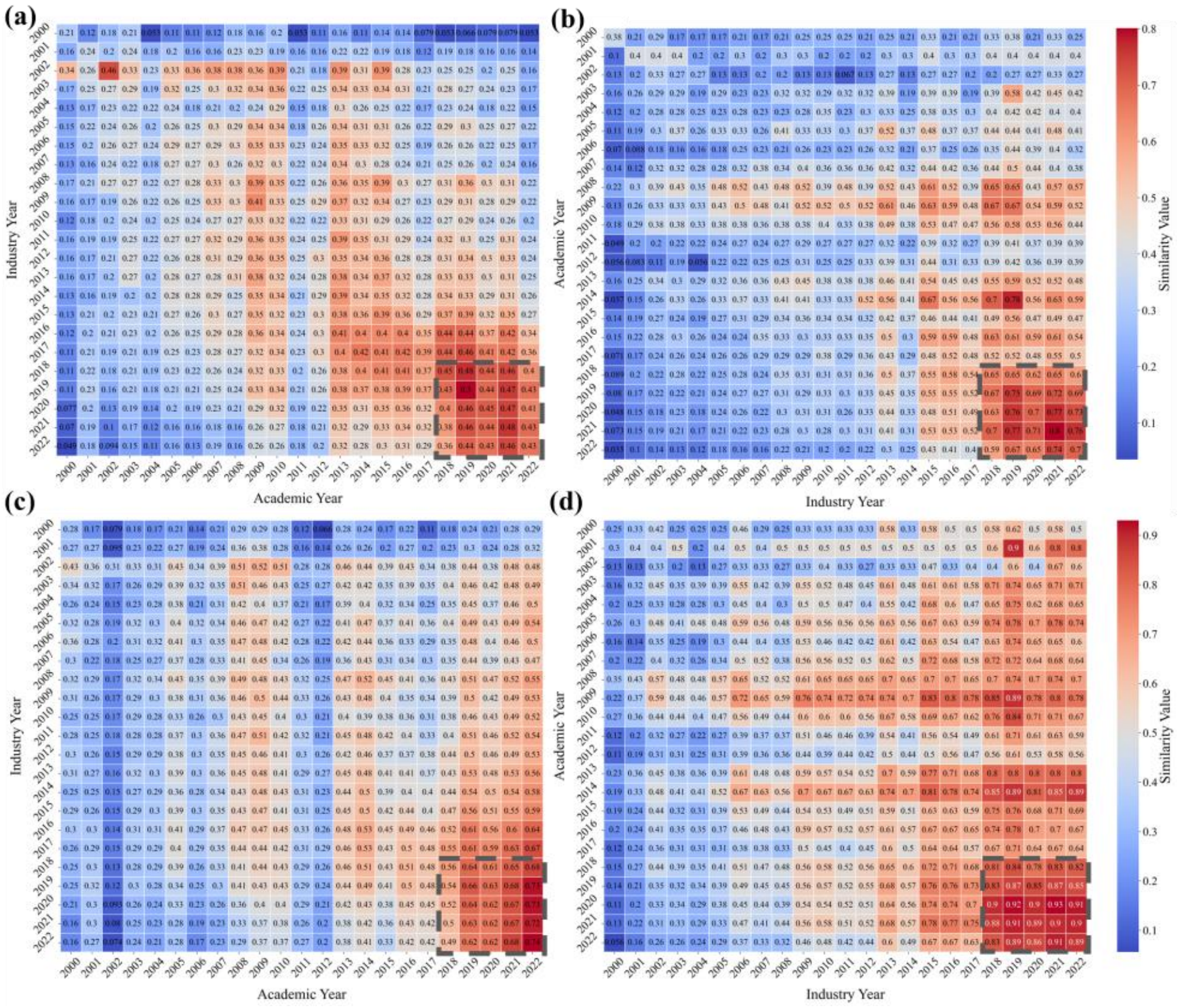

**Fig. B2.** Semantic similarity heatmaps (threshold: 5%/20%) (a) Academia searching for industry papers (5%); (b) Industry searching for academic papers (5%); (c) Academia searching for industry papers (20%); (d) Industry searching for academic papers (20%)

*Notes. The semantic similarity thresholds of 5% and 20% quantiles are 0.756 and 0.711, respectively*

**Fig. B3** shows the co-occurrence networks for each stage. **Fig. B3** (a) shows the core hubs of the academic network at the statistical machine learning stage as n-gram, Hidden Markov Model (HMM), Expectation-Maximization (EM), and Support Vector Machine (SVM), which constitute the statistical machine learning network. In addition, at the level of technology application, academic research also engaged in related studies, such as the study of "WordNet" semantic network (Miller, 1995), and the optimization of CRF feature engineering (Lafferty et al., 2001). In contrast, **Fig. B3** (b) shows that the industry focuses on the technology application path, building a technological closed loop around SMT (Statistical Machine Translation), such as the alignment template approach (Och & Ney, 2004), and focusing on specific applications.

In the deep learning stage, as shown in **Fig. B3** (c) and **Fig. B3** (d), the emerging pivot "Neural Network" as well as "RNN" and "LSTM" gradually replace the traditional methods, marking the significant improvement of sequence data processing capability. Notably, the Word2Vec model proposed by Mikolov et al. (2013) addressed the data sparsity issues of traditional discrete representations through efficient word vector space mapping, thereby establishing its classic status as a text vector modeling approach.

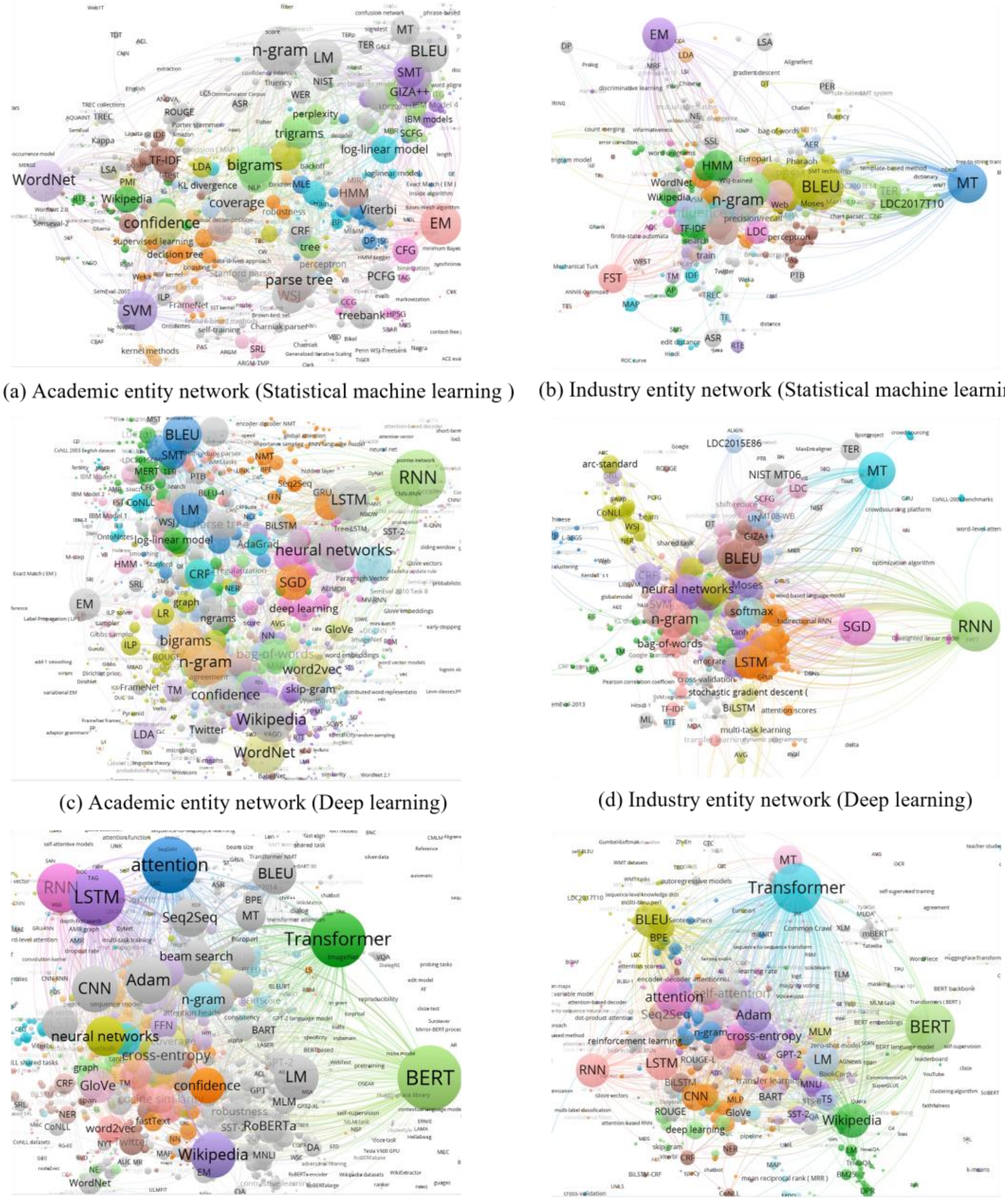

(a) Academic entity network (Statistical machine learning ) (b) Industry entity network (Statistical machine learning )

(c) Academic entity network (Deep learning) (d) Industry entity network (Deep learning)

(e) Academic entity network (Pre-training) (f) Industry entity network (Pre-training)

**Fig. B3.** Academic and industry entity networks in each period.

In the pre-training paradigm stage, the knowledge network topology has undergone a revolutionary transformation. Super-hubs such as “Transformer” (Vaswani et al., 2017) and “BERT” (Devlin et al., 2019) dominate the overall knowledge network. Under this new pre-training paradigm, both industry and academia have integrated knowledge resources, exhibiting a more similar trend in knowledge production.

# Appendix C. Reliability of indicators

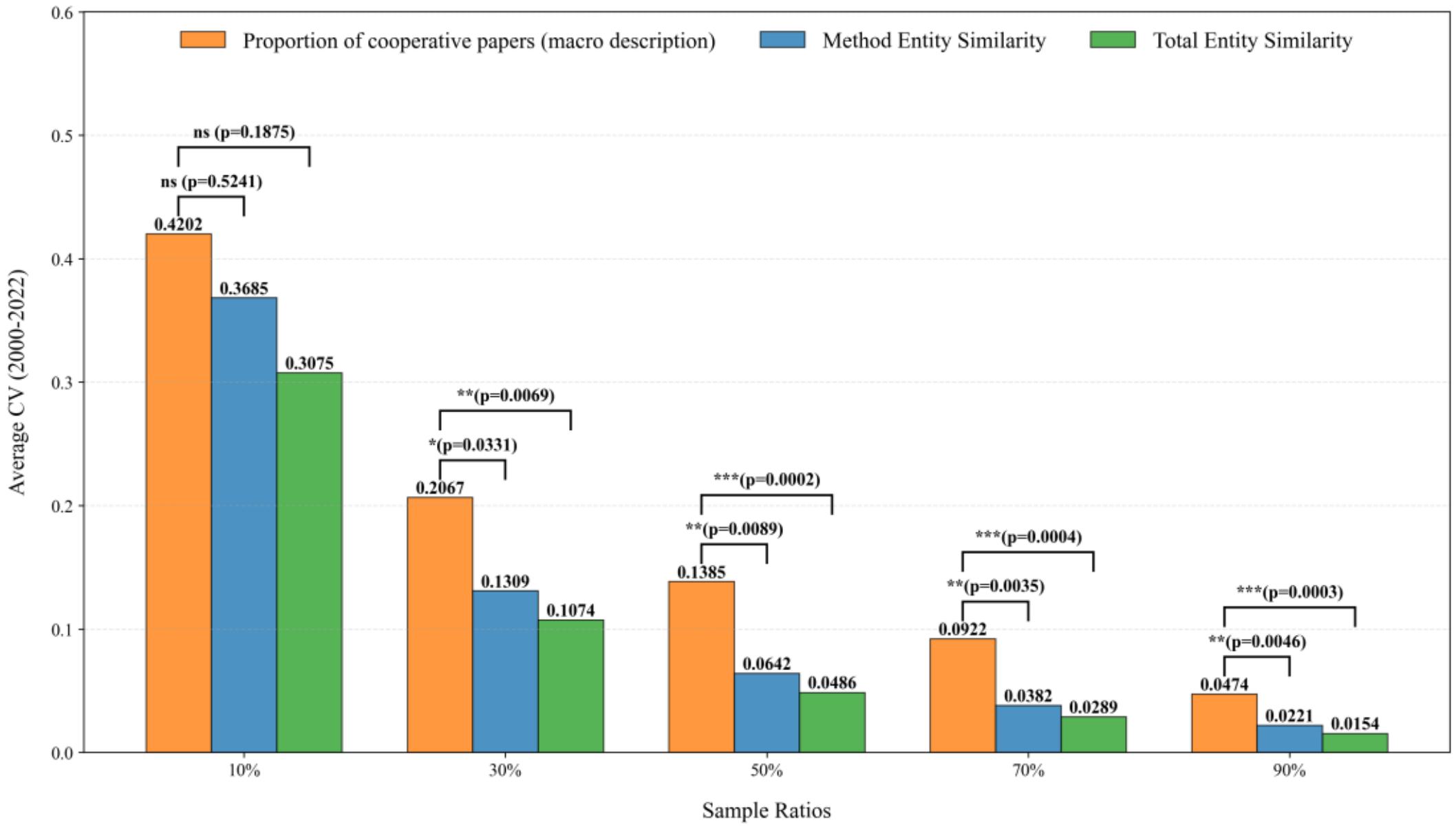

**Fig. C1.** Comparison of CV between indicators

*Notes.* ***: $p<0.001$, **: $p<0.01$, *: $p<0.05$.

To quantify the stability and reliability of different indicators, we use the CV as the evaluation metric. Essentially, the ratio of data standard deviation to mean, CV's key advantage, effectively eliminates the interference of dimension differences on stability assessment and intuitively reflects data dispersion. To control random errors, random sampling was conducted annually with proportions ranging from 10% to 90% at 20% intervals, and 10 parallel experiments were performed for each sampling proportion.

First, we calculated the annual CV:

$$CV_t(r) = \frac{\sigma_t(r)}{\mu_t(r)} \tag{C.1}$$

Where $\mu_t(r)$ is the mean of 10 experimental results in year $t$, and $\sigma_t(r)$ is the standard deviation of 10 experimental results in year $t$.

We measured the annual average CV to obtain the long-term stability of a certain indicator under a specific sampling ratio $r$.

$$\overline{CV}(r) = \frac{1}{23}\sum_{t=2000}^{2022} CV_t(r) \tag{C.2}$$

The experimental results are shown in **Fig. C1.** Both the similarity index based on all entities and the method entity similarity index have lower long-term stability than traditional macro indicators. To verify this difference's statistical significance, we used the Mann-Whitney U test. Results show that, except for the extreme scenario at a 10% sampling ratio, the proposed entity-based similarity indices' CV values are significantly lower than macro indicators, confirming that the proposed measurement method has higher stability and reliability.

# Appendix D. Summary of analytical dimensions and results

**Table D1. Summary of analytical dimensions, methods, and primary results**

| Analysis Dimensions | Methods | Primary Results | Chapter |
|---|---|---|---|
| Knowledge Entity Similarity | SciBERT | Similarity scores showed a gradual overall increase from 2000 to 2022, with a significant acceleration in growth following the pre-training phase in 2018. | 4.1.1 |
| Semantic similarity | SimCSE | The semantic similarity at critical technological inflection points exhibits a characteristic of sustained high similarity over the long term. | 4.1.2 |
| Academic Network Dominance | LCC | Academic leadership in the early stages, gradually moving toward balance | 4.2.1 |
| Number of overlapping nodes | Calculate shared nodes | Overlapping nodes gradually increase, with the most significant increase occurring during the pre-training phase | 4.2.2 |
| Network Modularity | Louvain algorithm | Modularity declines during the deep learning phase, while remaining stable at a low level during the pre-training phase. | 4.2.2 |
| Direction of Knowledge Flow | OpenAlex Search | From unidirectional to bidirectional | 4.3.1 |
| Knowledge Flow and Similarity Correlation | Correlation regression analysis | The strength of correlation varies across different stages. | 4.3.2 |
| Differences in Citation Behaviors | ECC; PSM | The ECC is higher in the industrial sector; Research with high computational demands cites academic literature less frequently. | 4.3.3 |

# Data availability

All the data and source code of this paper are freely available at the GitHub website: https://github.com/tinierZhao/Academic-Industrial-associations

## Acknowledgments

This paper was supported by the National Natural Science Foundation of China (Grant No.72074113)